\documentclass{article}

\usepackage{hyperref}
\usepackage{caption}
\usepackage[utf8]{inputenc} 
\usepackage[T1]{fontenc}    
\usepackage{url}            
\usepackage{booktabs}       
\usepackage{amsfonts}       
\usepackage{nicefrac}       
\usepackage{microtype}      
\usepackage{times}	   
\usepackage{adjustbox}		
\usepackage{array}		
\usepackage{multirow}           
\usepackage{tabularx}           
\usepackage{booktabs}           
\usepackage{rotating}
\usepackage{wrapfig}
\usepackage{nicefrac}
\usepackage{amsmath}
\usepackage{bm}
\usepackage{amsmath,amsthm,amssymb}
\usepackage{mathtools}
\usepackage{graphicx} 
\usepackage{subfigure} 
\usepackage[shortlabels]{enumitem}
\usepackage[ruled,vlined,algo2e,linesnumbered]{algorithm2e}

\newcommand{\vx}{{\bm{x}}}

\newcommand{\gauss}{\mbox{${\cal N}$}}

\newcommand{\argmin}{\operatornamewithlimits{arg\,min}}
\newcommand{\argmax}{\operatornamewithlimits{arg\,max}}
\newcommand{\ie}{i.\,e.}

\newcommand{\eg}{e.\,g.}

\newcommand{\hide}[1]{}

\newcommand{\ourmethod}{BOHB}
\newcommand{\fhcrc}[1]{\textcolor{red}{#1}}
\renewcommand{\fhcrc}[1]{#1}

\usepackage[accepted]{icml2018}

\icmltitlerunning{BOHB: Robust and Efficient Hyperparameter Optimization at Scale}

\begin{document}

\twocolumn[
  \icmltitle{BOHB: Robust and Efficient Hyperparameter Optimization at Scale}



\icmlsetsymbol{equal}{*}

\begin{icmlauthorlist}
\icmlauthor{Stefan Falkner}{ALUFR}
\icmlauthor{Aaron Klein}{ALUFR}
\icmlauthor{Frank Hutter}{ALUFR}
\end{icmlauthorlist}

\icmlaffiliation{ALUFR}{Department of Computer Science, University of Freiburg, Freiburg, Germany}

\icmlcorrespondingauthor{Stefan Falkner}{sfalkner@informatik.uni-freiburg.de}

\icmlkeywords{Machine Learning, ICML}

\vskip 0.3in
]



\printAffiliationsAndNotice{}  

\begin{abstract}
Modern deep learning methods are very sensitive to many hyperparameters, and, due to the long training times of state-of-the-art models, vanilla Bayesian hyperparameter optimization is typically computationally infeasible. On the other hand, bandit-based configuration evaluation approaches based on random search lack guidance and do not converge to the best configurations as quickly. Here, we propose to combine the benefits of both Bayesian optimization and bandit-based methods, in order to achieve the best of both worlds: strong anytime performance and fast convergence to optimal configurations. We propose a new practical state-of-the-art hyperparameter optimization method, which consistently outperforms both Bayesian optimization and Hyperband
on a wide range of problem types, including high-dimensional toy functions, support vector machines, feed-forward neural networks, Bayesian neural networks, deep reinforcement learning, and convolutional neural networks. Our method is robust and versatile, while at the same time being conceptually simple and easy to implement.
\end{abstract}

\section{Introduction}\label{sec:intro}

Machine learning has recently achieved great successes in a wide range of practical applications, but the performance of the most prominent methods depends more strongly than ever on the correct setting of many internal hyperparameters (see, e.g., \citet{henderson2017deep, melis2017state}).
The best-performing models for many modern applications of deep learning are getting ever larger and thus more computationally expensive to train, but at the same time both researchers and practitioners desire to set as many hyperparameters automatically as possible.
These constraints call for a practical solution to the hyperparameter optimization (HPO) problem that fulfills many desiderata:
\begin{description}[leftmargin=0cm]\itemsep-2pt
	\item[1. Strong Anytime Performance.] Since large contemporary neural networks often require days or even weeks to train, HPO methods that view performance as a black box function to be optimized require extreme resources.
	The overall budget that most researchers and practitioners can afford during development is often not much larger than that of fully training a handful of models, and hence practical HPO methods must go beyond this blackbox view to already yield good configurations with such a small budget.

	\item[2. Strong Final Performance.] On the other hand, what matters most at deployment time is the performance of the best configuration a HPO method can find given a larger budget. Since finding the best configurations in a large space requires guidance, this is where methods based on random search struggle.

	\item[3. Effective Use of Parallel Resources.] With the rise of parallel computing, large parallel resources are often available (e.g., compute clusters or cloud computing), and practical HPO methods need to be able to use these effectively. 

	\item[4. Scalability.] Modern deep neural networks require the setting of a multitude of hyperparameters, including architectural choices (e.g., the number and width
	of layers), optimization hyperparameters (e.g., learning rate schedules, momentum, and batch size), 
and regularization hyperparameters (e.g., weight decay and dropout rates). 
Practical modern HPO methods therefore must be able to easily handle problems ranging from just a few to many dozens of hyperparameters. 
	 
	\item[5. Robustness \& Flexibility.] The challenges for hyperparameter optimization vary substantially across subfields of machine learning; e.g., deep reinforcement learning systems are known to be very noisy~\cite{henderson2017deep}, while probabilistic deep learning is often very sensitive to a few key hyperparameters. Different hyperparameter optimization problems also give rise to different types of hyperparameters (such as binary, categorical, integer, and continuous), each of which needs to be handled effectively by a practical HPO method. 

\end{description}

\begin{figure}[t]
 \includegraphics[width=\columnwidth]{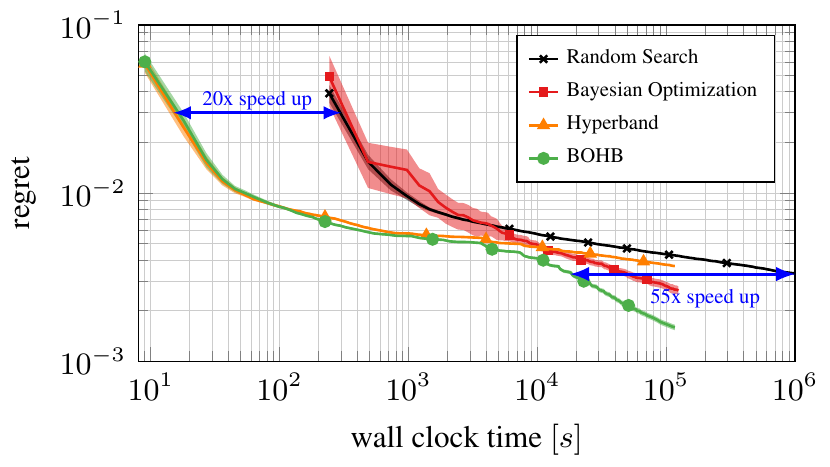}
\vspace*{-0.5cm}
\caption{Illustration of typical results obtained, here for optimizing six hyperparameters of a neural network. 
 We show the immediate regret of the best configuration found by 4 methods as a function of time. Hyperband has strong anytime performance, but for larger budgets does not perform much better than random search. In contrast, Bayesian optimization starts slowly (like random search), but given enough time outperforms Hyperband. Our new method BOHB achieves the best of both worlds, starting fast and also converging to the global optimum quickly.}
 \label{fig:teaser}
\end{figure}


As we will discuss in Section \ref{sec:related_work}, while there has been a lot of recent progress in the field of hyperparameter optimization, all existing methods have some strengths and weaknesses, but none of them fulfills all of these desiderata. The key contribution of this paper is therefore to combine the strengths of several methods (in particular, Hyperband~\cite{li-iclr17} and a robust \& effective variant~\cite{bergstra-nips11a} of Bayesian optimization~\cite{brochu-arXiv10a,shahriari-ieee16a}) to propose a practical HPO method that fulfills all of these desiderata. 
We first describe Bayesian optimization and Hyperband in more detail (Section \ref{sec:hyper_opt}) and then show how to combine them in our new method BOHB, as well as how to effectively parallelize the resulting system (Section \ref{sec:our_method}). 
Our extensive empirical evaluation (Section \ref{sec:experiments}) demonstrates that our method combines the best aspects of Bayesian optimization and Hyperband: it often finds good solutions over an order of magnitude faster than Bayesian optimization and converges to the best solutions orders of magnitudes faster than Hyperband. Figure \ref{fig:teaser} illustrates this pattern in a nutshell for optimizing six hyperparameters of a neural network. 



\section{Related Work on Model-based Hyperparamter Optimization}\label{sec:related_work}

\noindent \fhcrc{\textbf{Bayesian optimization} has been successfully applied to optimize hyperparameters of neural networks in many works:} \citet{snoek-nips12a} obtained state-of-the-art performance on CIFAR-10 by optimizing the hyperparameters of convolutional neural networks; \citet{bergstra-icml13a} used TPE \citep{bergstra-nips11a} to optimize a highly parameterized three layer convolutional neural network; and \citet{mendoza-automl16a} won 3 datasets in the 2016 AutoML challenge by automatically finding the right architecture and hyperparameters for fully-connected neural networks.

Gaussian processes are the most commonly-used probabilistic model in Bayesian optimization \citep{shahriari-ieee16a}, since they obtain smooth and well-calibrated uncertainty estimates.
However, Gaussian processes do not typically scale well to high dimensions and exhibit cubic complexity in the number of data points (scalability); they also do not apply to complex configuration spaces without special kernels (flexibility) and require carefully-set hyperpriors (robustness).

To speed up the hyperparameter optimization of machine learning algorithms, recent methods in Bayesian optimization try to extend the traditional blackbox setting by exploiting cheaper fidelities of the objective function \citep{swersky-archive14, klein-ejs17, swersky-nips13a, kandasamy2017multi, klein-iclr17, Poloczek2017}.
For instance, multi-task Bayesian optimization \citep{swersky-nips13a} exploits correlation between tasks to warm-start the optimization procedure.
Fabolas~\citep{klein-ejs17} uses similar techniques to evaluate configurations on subsets of the training data and to extrapolate their performance to the full dataset.
Even though these methods achieved both good anytime and final performance, they are based on Gaussian processes, which, as described above, do not satisfy all of our desiderata.
Alternative models, such as random forests~\cite{hutter-lion11a} or Bayesian neural networks \cite{snoek-icml15a,springenberg-nips2016,Perrone2017}, scale better with the number of dimensions, but with the exception of \citet{klein-iclr17} have not yet been adopted for multi-fidelity optimization. 

\noindent \textbf{Hyperband} \cite{li-iclr17} is a bandit strategy that dynamically allocates resources to a set of random configurations and uses successive halving \citep{jamieson-aistats16} to stop poorly performing configurations.
We describe this in more detail in Section \ref{sec:hyperband}.
Compared to Bayesian optimization methods that do not use multiple fidelities, Hyperband showed strong anytime performance, as well as flexibility and scalability to higher-dimensional spaces.
However, it only samples configurations randomly and does not learn from previously sampled configurations.
This can lead to a worse final performance than model-based approaches, as we show empirically in Section \ref{sec:experiments}.

Concurrently to our work, two other groups \cite{bertrandhyperparameter, wang-arxiv18} also attempted to combine Bayesian optimization with Hyperband. 
However, neither of them achieve the consistent and large speedups our method achieves. Furthermore, the method of \citet{bertrandhyperparameter} is based on  Gaussian processes and thus shares the limitations discussed above. \fhcrc{We discuss differences between our work and these two papers in more detail in Appendix B.} 

%

\section{Bayesian Optimization and Hyperband}\label{sec:hyper_opt}

The validation performance of machine learning algorithms can be modelled as a function $f: \mathcal{X} \rightarrow \mathbb{R}$ of their hyperparameters 
$\vx \in \mathcal{X}$.
We note that the hyperparameter configuration space $\mathcal{X}$ can include both discrete and continuous dimensions.
The hyperparameter optimization (HPO) problem is then defined as finding
$ \vx_{\star} \in \argmin_{\vx \in \mathcal{X}} f(\vx)$.

Due to the intrinsic randomness of most machine learning algorithms (\eg{} stochastic gradient descent), we assume that we cannot observe $f(\vx)$ directly but rather only trough noisy observations $y(\vx) = f(\vx) + \epsilon$, with $\epsilon \sim \gauss(0, \sigma^2_{noise})$.
We now discuss the two methods for tackling this optimization problem in more detail that we will use as components of our new method: Bayesian optimization and Hyperband.


\subsection{Bayesian Optimization}\label{sec:bo}

In each iteration $i$, Bayesian optimization (BO) uses a probabilistic model $p(f| D)$ to model the objective function $f$ based on the already observed data points $D = \{(\vx_0, y_0), \dots, (\vx_{i-1}, y_{i-1})\}$.
BO uses an acquisition function $a: \mathcal{X} \rightarrow \mathbb{R}$ based on the current model $p(f | D)$ that trades off exploration and exploitation.
Based on the model and the acquisition function, it iterates the following three steps:
(1)~select the point that maximizes the acquisition function $\vx_{new} = \argmax_{\vx \in \mathcal{X}} a(\vx)$, (2)~evaluate the objective function $y_{new} = f(\vx_{new}) + \epsilon$, and (3)~augment the data $D \leftarrow D \cup  (\vx_{new}, y_{new})$ and refit the model.
A common acquisition function is the expected improvement (EI) over the currently best observed value $\alpha = \min \{y_0, \ldots, y_n \}$:
\begin{equation}\label{eq:ei}
  a(\vx) = \int \max(0, \alpha - f(\vx))dp(f|D).
\end{equation}

\paragraph{Tree Parzen Estimator.}
The Tree Parzen Estimator (TPE) \cite{bergstra-nips11a} is a Bayesian optimization method that uses a kernel density estimator to model the densities
\begin{equation}
	\begin{aligned}
		l(\vx) &= p(y < \alpha | \vx, D) \\ 
		g(\vx) &= p(y > \alpha | \vx, D)
	\end{aligned}
	\label{eq:tpe_densities}
\end{equation}
over the input configuration space instead of modeling the objective function $f$ directly by $p(f|D)$.
To select a new candidate $\vx_{new}$ to evaluate, it maximizes the ratio $\nicefrac{l(\vx)}{g(\vx)}$; 
\citet{bergstra-nips11a} showed that this is equivalent to maximizing EI in Equation \eqref{eq:ei}.
Due to the nature of kernel density estimators, TPE easily supports mixed continuous and discrete spaces, and model construction scales linearly in the number of data points (in contrast to the cubic-time Gaussian processes (GPs) predominant in the BO literature). 

\subsection{Hyperband}\label{sec:hyperband}

While the objective function $f: \mathcal{X} \rightarrow \mathbb{R}$ is typically expensive to evaluate (since it requires training a machine learning model with the specified hyperparameters), in most applications it is possible to define cheap-to-evaluate approximate versions $\tilde{f}(\cdot,b)$ of $f(\cdot)$ that are parameterized by a so-called \emph{budget} $b \in [b_{min}, b_{max}]$. With the maximum budget $b=b_{max}$, we have $\tilde{f}(\cdot,b_{max}) = f(\cdot)$, whereas with $b<b_{max}$, $\tilde{f}(\cdot,b)$ is only an approximation of $f(\cdot)$ whose quality typically increases with $b$. In our experiments, we will use this budget to encode the number of iterations for an iterative algorithm, the number of data points used, the number of steps in an MCMC chain, and the number of trials in deep reinforcement learning.

Hyperband (HB) \cite{li-iclr17} is a multi-armed bandit strategy for hyperparameter optimization that takes advantage of these different budgets $b$ by repeatedly calling SuccessiveHalving (SH) \cite{jamieson-aistats16} to identify the best out of $n$ randomly sampled configurations.
It balances very agressive evaluations with many configurations on the smallest budget, and very conservative runs that are directly evaluated on $b_{max}$.
The exact procedure for this trade-off is shown in Algorithm \ref{alg:HB} (with pseudocode for SH shown in Appendix C).
Line 1 computes the geometrically spaced budget $ \in [b_{min}, b_{max}]$.
The number of configurations sampled in line 3 is chosen such that every SH run requires the same total budget.
SH internally evaluates configurations on a given budget, ranks them by their performance, and continues the top $\eta^{-1}$ (usually the best-performing third) on a budget $\eta$ times larger.
This is repeated until the maximum budget is reached.
In practice, HB works very well and typically outperforms random search and Bayesian optimization methods operating on the full function evaluation budget quite easily for small to medium total budgets.
However, its convergence to the global optimum is limited by its reliance on randomly-drawn configurations, and with large budgets its advantage over random search typically diminishes.

\begin{algorithm2e}[t]
	\caption{Pseudocode for Hyperband using SuccessiveHalving (SH) as a subroutine.}\label{alg:HB}
        \DontPrintSemicolon
        \SetKwInOut{Input}{input}\SetKwInOut{Output}{output}
	\Input{budgets $b_{min}$ and $b_{max}$, $\eta$}
	$s_{max} = \lfloor \log_{\eta} \frac{b_{max}}{b_{min}} \rfloor $\;

	\For{$s \in \{ s_{max}, s_{max}-1,\dots,0 \}$}{
		sample $n = \lceil \frac{s_{max}+1}{s+1}\cdot \eta^s \rceil$\ configurations\;
		run SH on them with $\eta^s \cdot b_{max}$ as initial budget
	}
\end{algorithm2e}

\section{Model-Based Hyperband}\label{sec:our_method}


We now introduce our new practical HPO method, which we dub \emph{\ourmethod{}} since it combines Bayesian optimization (BO) and Hyperband (HB).
We designed \ourmethod{} to satisfy all the desiderata described in the introduction. HB already satisfies most of these desiderata (in particular, strong anytime performance, scalability, robustness and flexibility), and we combine it with BO to also satisfy the desideratum of strong final performance in \ourmethod{}. We also describe how to extend \ourmethod{} to make effective use of parallel resources.  

In the design of \ourmethod{}'s BO component, on top of the five desiderata above, we also followed two additional ones: 
\begin{description}[leftmargin=0cm]\itemsep-2pt
  \item[6. Simplicity.] Simplicity is a virtue, since simple approaches can be easily verified, have less components that can break, and can be easily reimplemented in different frameworks. HB is very simple, but standard GP-BO methods are not: they tend to require complex approximations, complex MCMC sampling over hyperparameters, and for good performance also data-dependent choices of kernel functions and hyperpriors. 
  \item[7. Computational efficiency.] Since our HB component allows us to carry out many function evaluations at small budgets, the cubic complexity of standard GPs, and even the lower complexity of approximate GPs would become problematic. Furthermore, compared to these cheap function evaluations, the complexity of computing sophisticated acquisition functions may also become a bottleneck, especially when parallelization effectively reduces the cost of function evaluations.
\end{description}
For these reasons, along with the reasons of scalability, robustness \& flexibility, we based \ourmethod{}'s BO component on the simple TPE method discussed above. As reliable GP-based BO methods become available that satisfy all the desiderata above, it would be easy to replace TPE with them.

\subsection{Algorithm description}
BOHB relies on HB to determine how many configurations to evaluate with which budget, but it replaces the random selection of configurations at the beginning of each HB iteration by a model-based search. Once the desired number of configurations for the iteration is reached, the standard successive halving procedure is carried out using these configurations. We keep track of the performance of all function evaluations $g(\vx,b) + \epsilon$ of configurations $\vx$ on all budgets $b$ to use as a basis for our models in later iterations.

We follow HB's way of choosing the budgets and continue to use SH, but we replace the random sampling by a BO component to guide the search.
We construct a model and use BO to select a new configuration, based on the configurations evaluated so far.
In the remainder of this section, we will explain this procedure summarized by the pseudocode in Algorithm \ref{alg}.

The BO part of BOHB closely resembles TPE, with one major difference: we opted for a single multidimensional KDE compared to the hierarchy of one-dimensional KDEs used in TPE
in order to better handle interaction effects in the input space.
To fit useful KDEs (in line 4 of Algorithm \ref{alg}),
we require a minimum number of data points $N_{min}$; this is set to $d + 1$ for our experiments, where $d$ is the number of hyperparameters.
To build a model as early as possible, we do not wait until $N_b = \vert D_b \vert$, the number of observations for budget $b$, is large enough to satisfy $q \cdot N_b \geq N_{min}$.
Instead, after initializing with $N_{min}+2$ random configurations (line 3), we choose the 
\begin{equation}
\begin{aligned}
	N_{b, l} &= \max(N_{min}, q\cdot N_b)\\
	N_{b, g} &= \max(N_{min}, N_b - N_{b, l})
\end{aligned}
\label{eq:kde_split}
\end{equation}
best and worst configurations, respectively, to model the two densities.
This ensures that both models have enough datapoints and have the least overlap when only a limited number of observations is available.
We used the KDE implementation from statsmodels~\citep{seabold2010statsmodels}, estimating the KDE's bandwidth with the default estimation procedure (Scott's rule of thumb), which is efficient and performed well in our experience. Details on our KDE are given in Appendix D.

As the optimization progresses, more configurations are evaluated on bigger budgets.
Given that the goal is to optimize on the largest budget, {\ourmethod} always uses the model for the largest budget for which enough observations are available (line 2).
This enables it to overcome wrong conclusions drawn on smaller budgets by eventually relying on results with the highest fidelity only.

To optimize EI (lines 5-6), we sample $N_s$ points from $l'(\vx)$, which is the same KDE as $l(\vx)$ but with all bandwidths multiplied by a factor $b_w$ to encourage more exploration around the promising configurations.
We observed that this improves convergence especially in the late stages of the optimization, when the model on the biggest budget is queried frequently but updated rarely.


In order to keep the theoretical guarantees of HB, we also sample a constant fraction $\rho$ of the configurations uniformly at random (line 1).
Besides global exploration, this guarantees that after $m\cdot(s_{max}+1)$ SH runs, our method has (on average) evaluated $\rho\cdot m \cdot (s_{max}+1)$ random configurations on $b_{max}$.
As every SH run consumes a budget of at most $(s_{max}+1)\cdot b_{max}$, in the same time random search evaluates $(\rho^{-1}\cdot (s_{max}+1))$-times as many configuration on the largest budget.
This means, that in the worst case (when the lower fidelities are misleading), {\ourmethod} is at most this factor times slower than RS, but it is still guaranteed to converge eventually.
The same argument holds for HB, but in practice both HB and \ourmethod{} substantially outperform RS in our experiments.

\begin{algorithm2e}[b]
\caption{Pseudocode for sampling in {\ourmethod}}\label{alg}
        \DontPrintSemicolon
        \SetKwInOut{Input}{input}\SetKwInOut{Output}{output}
	\Input{observations $D$, fraction of random runs $\rho$, percentile $q$, number of samples $N_s$, minimum number of points $N_{min}$ to build a model, and bandwidth factor $b_w$}
        \Output{next configuration to evaluate}

        \lIf{rand() < $\rho$}{\Return random configuration}

	$b = \argmax \left\{ D_b:  \vert D_b \vert \geq N_{min} + 2\right\}$\; 
        \lIf{ $b=\emptyset$}{ \Return random configuration}

	fit KDEs according to Eqs.~\eqref{eq:tpe_densities} and ~\eqref{eq:kde_split}\;
	draw $N_s$ samples according to $l'(\vx)$ (see text)\;
        \Return sample with highest ratio  $l(\vx)/g(\vx)$
\end{algorithm2e}

No optimizer is free of hyperparameters itself, and their effects have to be studied carefully.
We therefore include a detailed empirical analysis of \ourmethod{}'s hyperparameters in Appendix G that
shows each hyperparameter's effect when all others are fixed to their default values (these are also listed there).
We find that {\ourmethod} is quite insensitive to its hyperparameters, with the default working robustly across different scenarios.

\subsection{Parallelization}
Modern optimizers must be able to take advantage of parallel resources effectively and efficiently.
{\ourmethod} achieves that by inheriting properties from both TPE and HB.
The parallelism in TPE is achieved by limiting the number of samples to optimize EI, purposefully not optimizing it fully to obtain diversity.
This ensures that consecutive suggestions by the model are diverse enough to yield near-linear speedups when evaluationed in parallel. 
On the other hand, HB can be parallelized by (a) starting different iterations at the same time (a parallel for loop in Alg. \ref{alg:HB}), and (b) evaluating configurations concurrently within each SH run.

Our parallelization strategy of {\ourmethod} is as follows.
We start with the first SH run that sequential HB would perform (the most aggressive one, starting from the lowest budget), sampling configurations with the strategy outlined in Algorithm \ref{alg} until either (a) all workers are busy, or (b) enough configurations have been sampled for this SH run.
In case (a), we simply wait for a worker to free up and then sample a new configuration. 
In case (b), we start the next SH run in parallel, sampling the configurations to run for it also according to Algorithm \ref{alg}; observations $D$ (and therefore the resulting models) are shared across all SH runs.
\ourmethod{} is an anytime algorithm that at each point in time keeps track of the configuration that achieved the best validation performance; it can also be given a maximum budget of SH runs.  

\begin{figure}[t]
	\centering
\vspace*{-0.5cm}
	\includegraphics[width=\columnwidth]{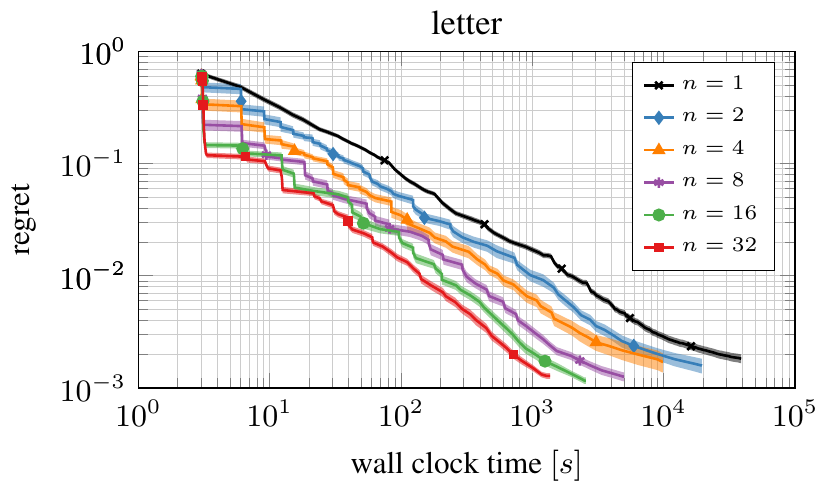}
	\caption{Performance of our method with different number of parallel workers on the letter surrogate benchmark (see Sec. \ref{sec:experiments}) for 128 iterations. The speedup for two and four workers is close to linear, for more workers it becomes sublinear. For example, the speedup to achieve a regret of $10^{-2}$ for one vs. 32 workers is ca. $2000s / 130s \approx 15$. We plot the mean and twice the standard error of the mean over 128 runs.}
	\label{fig:parallel_letter}
	\vspace{-0.1cm}
\end{figure}

We note that SH has also been parallelized in (so far unpublished) independent work~\citep{li2018massively}.
Next to parallelizing SH runs (by filling the next free worker with the ready-to-be-executed run with the largest budget), that work mentioned that HB can trivially be parallelized by running its SH runs in parallel.
In contrast to this approach of parallelizing HB by having separate pools of workers for each SH run, we rather join all workers into a single pool, and whenever a worker becomes available preferentially execute waiting runs with smaller budgets.
New SH runs are only started when the SH runs currently executed are not waiting for a worker to free up.
This strategy (a)~allows us to achieve better speedups by using all workers in the most aggressive (and often most effective) bracket first, and (b)~also takes full advantage of models built on smaller budgets.
Figure \ref{fig:parallel_letter} demonstrates that our method of parallelization can effectively exploit many parallel workers.


\section{Experiments}\label{sec:experiments}

We now comprehensively evaluate \ourmethod{}'s empirical performance in a wide range of tasks, including a high-dimensional toy function, as well as optimizing the hyperparameters of support vector machines, feed-forward neural networks, Bayesian neural networks, deep reinforcement learning agents and convolutional neural networks.
%
Code for \ourmethod{} and our benchmarks is publicly available at \href{https://github.com/automl/HpBandSter}{https://github.com/automl/HpBandSter}

To compare against TPE, we used the Hyperopt package \cite{bergstra-nips11a}, and for all GP-BO methods we used the RoBO python package \cite{klein-bayesopt17}.
In all experiments we set $\eta = 3$ for HB and \ourmethod{} as recommended by \citet{li-iclr17}.
If not stated otherwise, for all methods we report the mean performance and the standard error of the mean of the best observed configuration so far (incumbent) at a given budget.

\subsection{Artificial Toy Function: Counting Ones}

In this experiment we investigated \ourmethod{}'s behavior in high-dimensional mixed continuous / categorical configuration spaces.
Since GP-BO methods do not work well on such configuration spaces \cite{eggensperger-bayesopt13} we do not include them in this experiment.
However, we do use SMAC \cite{hutter-lion11a}, since its random forest are known to perform well in high-dimensional categorical spaces \cite{eggensperger-bayesopt13}.

Given a set of $N_{cat}$ categorical variables $x \in \{0, 1\}$ and $N_{cont}$ continuous variables $x \in [0, 1]$, we defined the counting one problem as:
\vspace*{-0.2cm}
\[
  f(\vx) = - ( \sum_{i=0}^{N_{cat}} x_i + \sum_{j=N_{cat} + 1}^{N_{cat} + N_{cont}} \mathbb{E}_{X \sim B_j(X)} [ X ]).
\]
\vspace*{-0.4cm}

The expectation is taken with respect to a Bernoulli distribution $B_j$ with parameter $p = x_j$.
As a budget for HB and \ourmethod{}, we used the number of samples $b \in [9, 729]$ allowed to approximate this expectation; all other methods always evaluated on the full budget, \ie{}, $b=729$.

\begin{figure}[t]
 \includegraphics[width=\columnwidth]{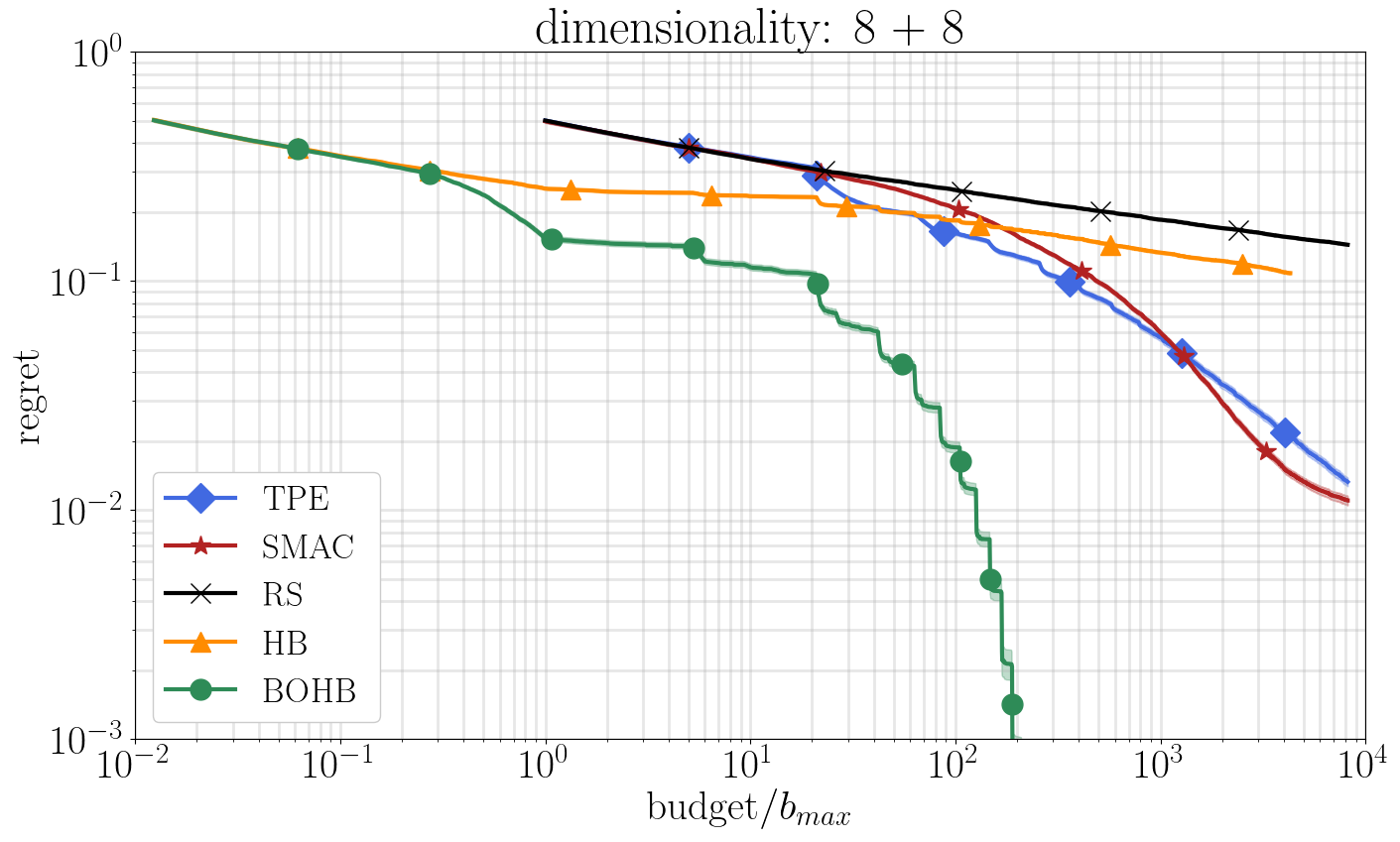}
\vspace*{-0.4cm} 
\caption{Results for the counting ones problem in 16 dimensional space with 8 categorical and 8 continuous hyperparameters. In higher dimensional spaces RS-based methods need exponentially more samples to find good solutions.}
 \label{fig:counting_ones}
\vspace{-0.4cm}
\end{figure}

For each method, we performed 512 independent runs and report the immediate regret $\vert f(\vx_{inc}) - f(\vx_*) \vert$ where $\vx_* \in \argmin f(\vx)$ and $\vx_{inc}$ is the incumbent at a specific time step.
Figure \ref{fig:counting_ones} shows the results for a 16-dimensional space with $N_{cat}=8$ and $N_{cont}=8$ parameters.
The results for other dimensions can be found in Appendix H.

Random search worked very poorly on this benchmark and was quickly dominated by the model-based methods SMAC and TPE.
Even though HB was faster in the beginning, SMAC and TPE clearly outperformed it after having obtained a sufficiently informative model. 
\ourmethod{} worked as well as HB in the beginning and then quickly started to perform better and --- as the only method --- converged in the time budget.
We obtained similar results for other dimensionalities (see Figure 7 in the supplementary material). However, we note that with as many as 64 dimensions, TPE and SMAC
started to perform better than \ourmethod{} since the noise grows and evaluating configurations on a smaller budget does not help to build better models for the full budget.
 

\subsection{Comprehensive Experiments on Surrogate Benchmarks}

For the next experiments we constructed a set of surrogate benchmarks based on offline data following \citet{eggensperger-aaai15}.
Optimizing a surrogate instead of the real objective function is substantially cheaper, which allows us to afford many independent runs for each optimizer and to draw statistically more meaningful conclusions.
A more detailed discussion of how we generated these surrogates can be found in Appendix I in the supplementary material.
To better compare the convergence towards the true optimum, we again computed the immediate regret of the incumbent.

\subsubsection{Support Vector Machine on MNIST}

To compare against GP-BO, we used the support vector machine on MNIST surrogate from \citet{klein-ejs17}.
This surrogate imitates the hyperparameter optimization of a support vector machine with a RBF kernel with two hyperparameters: the regularization parameter $C$ and the kernel parameter $\gamma$.
The budget is given by the number of training datapoints, where the minimum budget is $\nicefrac{1}{512}$ of the training data and the maximum budget is the full training data.
For further details, we refer to \citet{klein-ejs17}.

\begin{figure}[t]
  \includegraphics[width=\columnwidth]{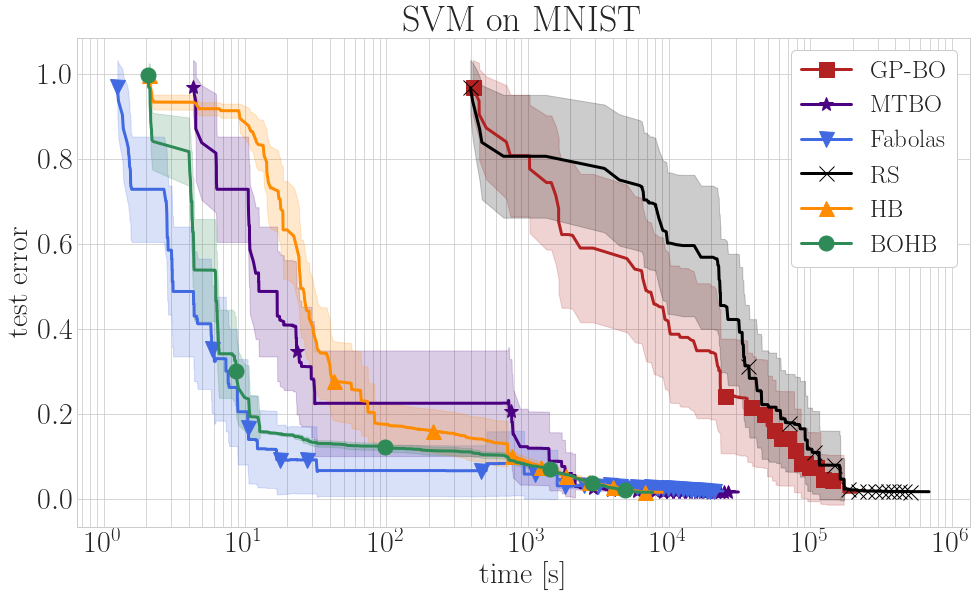}
\vspace*{-0.5cm} 
 \caption{Comparison on the SVM on MNIST surrogates as described in \citet{klein-ejs17}. \ourmethod{} works similarly to Fabolas on this two dimensional benchmark and outperforms MTBO and HB.}
 \label{fig:svm_mnist}
  \vspace{-0.4cm}
\end{figure}

Figure \ref{fig:svm_mnist} compares \ourmethod{} to various BO methods, such as Fabolas \cite{klein-ejs17}, multi-task Bayesian optimization (MTBO) \cite{swersky-nips13a}, GP-BO with expected improvement \cite{snoek-nips12a, klein-bayesopt17}, RS and HB.
\ourmethod{} achieved similar performance as Fabolas and worked slightly better than HB.
We note that this is a low-dimensional continuous problem, for which it is well known that GP-BO methods usually work better than other methods, such as kernel density estimators \citep{eggensperger-bayesopt13}.

\subsubsection{Feed-forward Neural Networks on OpenML Datasets}

We optimized six hyperparameters that control the training procedure (initial learning rate, batch size, dropout, exponential decay factor for learning rate) and the architecture (number of layers, units per layer) of a feed forward neural network for six different datasets gathered from OpenML \citep{vanschoren-sigkdd13a}: Adult~\citep{kohavi1996scaling},  Higgs~\citep{baldi2014searching}, Letter~\citep{Frey1991}, MNIST~\citep{lecun-isp01a}, Optdigits~\citep{lichman-13}, and Poker~\citep{cattral2002evolutionary}.
A detailed description of all hyperparameter ranges and training budgets can be found in Appendix I.

We ran random search (RS), TPE, HB, GP-BO, Hyperband with LC-Net (HB-LCNet, see~\citet{klein-iclr17}) and {\ourmethod} on all six datasets and summarize the results for one of them in Figure \ref{fig:paramnet_methods_subset}.
Figures for the other datasets are shown in Appendix E.

We note that HB initially performed much better than the vanilla BO methods and achieved a roughly three-fold speedup over RS.
However, for large enough budgets TPE and GP-BO caught up in all cases, and in the end found better configurations than HB and RS. 
HB and \ourmethod{} started out identically, but \ourmethod{} achieved the same final performance as HB 100 times faster, while at the same time yielding a final result that was better than that of the other BO methods.
All model-based methods substantially outperformed RS at the end of their budget, whereas HB approached the same performance.
Interestingly, the speedups that TPE and GP-BO achieved over RS are comparable to the speedups that {\ourmethod} achieved over HB.
Finally, HB-LCNet performed somewhat better than HB alone, but consistently worse than \ourmethod{}, even when tuning HB-LCNet. We only compare to HB-LCNet on this benchmark, since it is the only one that includes full learning curves (for which the parametric functions in HB-LCNet were designed). Also, HB-LCNet requires access to performance values for all budgets, which we do not obtain when, e.g., using data subset sizes as a budget, and we thus expect HB-LCNet to perform poorly in the other cases.

\begin{figure}[t]
	\centering
	\includegraphics[width=\columnwidth]{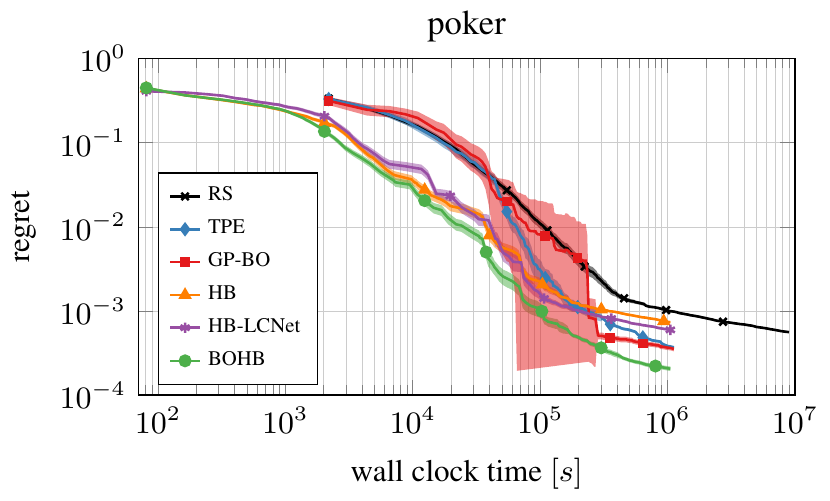}
\vspace*{-0.7cm}	
\caption{Optimizing six hyperparameter of a feed-forward neural network on featurized datasets; results are based on surrogate benchmarks. Results for the other 5 datasets are qualitatively similar and are shown in Figure 1 in the supplementary material.}
	\label{fig:paramnet_methods_subset}
\vspace*{-0.3cm}	
\end{figure}

\subsection{Bayesian Neural Networks}

For this experiment we optimized the hyperparameters and the architecture of a two-layer fully connected Bayesian neural network trained with Markov Chain Monte-Carlo (MCMC) sampling.
We used stochastic gradient Hamiltonian Monte-Carlo sampling (SGHMC) \cite{chen-icml14} with scale adaption \cite{springenberg-nips2016} to sample the parameter vector of the network.
Note that to the best of our knowledge this is the first application of hyperparameter optimization for Bayesian neural networks.

As tunable hyperparameters, we exposed the step length, the length of the burn-in period, the number of units in each layer, and the decay parameter of the momentum variable. 
A detailed description of the configuration space can be found in Appendix J.
We used the Bayesian neural network implementation provided in the RoBO python package \cite{klein-bayesopt17} as described by \citet{springenberg-nips2016}.

We considered two different UCI \cite{lichman-13} regression datasets, \textit{Boston housing} and \textit{protein structure} as described by \citet{lobato-icml15} and report the negative log-likelihood of the validation data.
For \ourmethod{} and HB, we set the minimum budget to $500$ MCMC steps and the maximum budget to $10000$ steps.
RS and TPE evaluated each configuration on the maximum budget.
For each hyperparameter optimization method, we performed 50 independent runs to obtain statistically significant results.

\begin{figure}[t]
 \centering
  \includegraphics[width=0.49\textwidth]{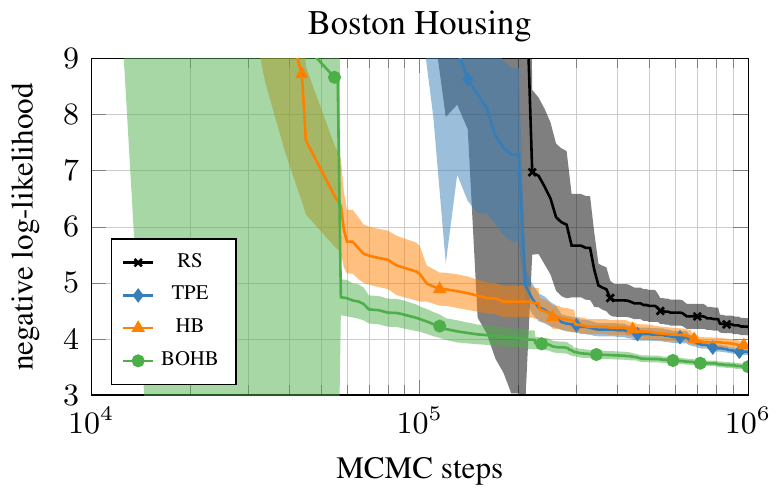}
\vspace*{-0.4cm}
  \caption{Optimization of 5 hyperparameters of a Bayesian neural network trained with SGHMC. Many random hyperparameter configurations lead to negative log-likelihoods orders of magnitude higher than the best performing ones. We clip the y-axis at 9 to ensure visibility in the plot.}
\label{fig:bnn}
\vspace*{-0.2cm}
\end{figure}

As Figure \ref{fig:bnn} shows, HB initially performed better than TPE, but TPE caught up given enough time.
\ourmethod{} converged faster than both HB and TPE and even found a better configuration than the baselines on the Boston housing dataset.

\subsection{Reinforcement Learning}

Next, we optimized eight hyperparameters of proximal policy optimization (PPO) \cite{schulman2017proximal} to learn the \textit{cartpole swing-up} task.
For PPO, we used the implementation from the TensorForce framework developed by \citet{schaarschmidt2017tensorforce} and we used the implementation from OpenAI Gym \cite{brockman-corr16} for the cartpole environment.
The configuration space for this experiment can be found in Appendix K.

To find a configuration that not only converges quickly but also works robustly, for each function evaluation we ran a configuration for nine individual trials with a different seed for the random number generator.
We returned the average number of episodes until PPO has converged to the optimum, defining convergence to mean that the reinforcement learning agent achieved the highest possible reward for 20 consecutive episodes.
For each hyperparameter configuration we stopped training after the agent has either converged or ran for a maximum of 3000 episodes.
The minimum budget for \ourmethod{} and HB was one trial and the maximum budget were nine trials, and all other methods used a fixed number of nine trials.
As in the previous benchmark, for each hyperparameter optimization method we performed 50 independent runs.

\begin{figure}[t]
\includegraphics[width=\columnwidth]{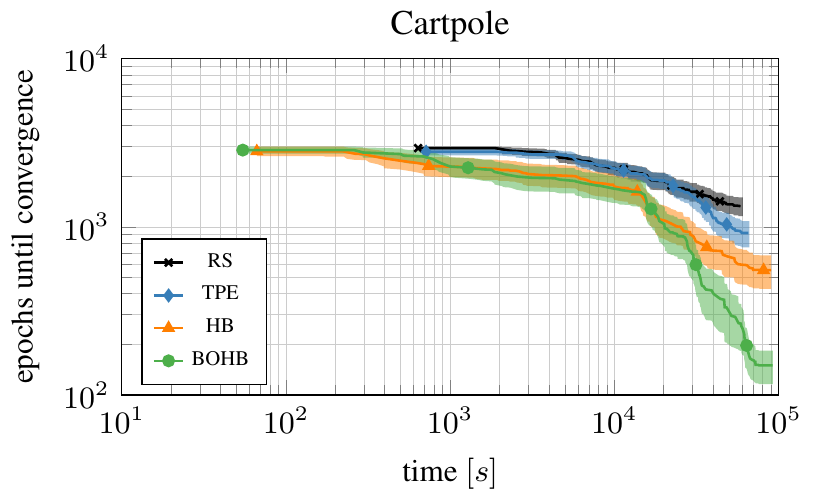}
\vspace*{-0.4cm}
 \caption{Hyperparameter optimization of 8 hyperparameters of PPO on the cartpole task. \ourmethod{} starts as well as HB but converges to a much better configuration.}
 \label{fig:cartpole}
\end{figure}

Figure \ref{fig:cartpole} shows that HB and \ourmethod{} worked equally well in the beginning, but \ourmethod{} converged to better configurations in the end.
Apparently, the budget for this benchmark was not sufficient for TPE to find the same configuration.

\subsection{Convolutional Neural Networks on CIFAR-10}
For a final evaluation, we optimized the hyperparameters of a medium-sized residual network (depth 20 and basewidth of 64; roughly $8.5\mathrm{M}$ parameters) with Shake-Shake \citep{gastaldi2017shake} and Cutout \citep{devries2017improved} regularization. To perform hyperparameter optimization, we split off 5\,000 training images as a validation set.
As hyperparameters, we optimized learning rate, momentum, weight decay, and batch size. 

We ran {\ourmethod} with budgets of 22, 66, 200, and 600 epochs, using 19 parallel workers.
Each worker used 2 NVIDIA TI 1080 GPUs for parallel training, which resulted in runs with the longest budget taking approximately $7$ hours (on 2 GPUs).
The complete \ourmethod{} run of 16 iterations required a total of 33 GPU days (corresponding to a cost of less than 3 full function evaluations on each of the 19 workers) and achieved a test error of $2.78\% \pm 0.09\%$ (which is better than the error 
\citet{gastaldi2017shake} obtained with a slightly larger network).
%
%
%
While we note that the performance numbers from different papers are not directly comparable due to the use of different optimization and regularization approaches, it is still instructive to compare this result to others in the literature.
Our result is better than that of last year's state-of-the-art neural architecture search by reinforcement learning (3.65\%~\cite{zoph-iclr17a}) and the recent paper on progressive neural architecture search (3.41\%~\cite{progressive}), but it does not quite reach the state-of-the-art performance of 2.4\% and 2.1\% reported in recent arXiv papers on reinforcement learning~\cite{zoph_arxiv2017} and evolutionary search~\cite{real_regularized_2018}. However, since these approaches used 60 to 95 times more compute resources (2\,000 and 3\,150 GPU days, respectively!), as well as networks with 3-4 more parameters, we believe that our results are a strong indication of the practical usefulness of \ourmethod{} for resource-constrained optimization. 
%

%

\section{Conclusions}\label{sec:conclusions}
We introduced {\ourmethod}, a simple yet effective method for hyperparameter optimization satisfying the desiderata outlined above: it is robust, flexible, scalable (to both high dimensions and parallel resources), and achieves both strong anytime performance and strong final performance.
We thoroughly evaluated its performance on a diverse set of benchmarks and demonstrated its improved performance compared to a wide range of other state-of-the-art approaches.
Our easy-to-use open-source implementation (available under \href{https://github.com/automl/HpBandSter}{https://github.com/automl/HpBandSter}) should allow the community to effectively use our method on new problems.
To further improve {\ourmethod}, we will consider an automatic adaptation of the budgets used to alleviate the problem of misspecification by the user while maintaining the versatility and robustness of the current version.

\section*{Acknowledgements}
We thank Ilya Loshchilov for suggesting to track the best hyperparameter setting across different budgets (already in late 2015), which influenced our thoughts about the problem and ultimately the development of \ourmethod{}.   
This work has partly been supported by the European Research Council (ERC) under the European Union's Horizon 2020 research and innovation programme under grant no. 716721, by the European Commission under grant no. H2020-ICT-645403-ROBDREAM, and by the German Research Foundation (DFG) under Priority Programme Autonomous Learning (SPP 1527, grant BR 3815/8-1 and HU 1900/3-1)
Furthermore, the authors acknowledge support by the state of Baden-Württemberg through bwHPC and the DFG through grant no INST 39/963-1 FUGG.

\begin{footnotesize}
\bibliography{local,strings,lib,proc}
\bibliographystyle{icml2018}
\end{footnotesize}
\end{document}


\twocolumn[
  \icmltitle{Supplementary material for:\\ BO-HB: Robust and Efficient Hyperparameter Optimization at Scale}



\icmlsetsymbol{equal}{*}

\begin{icmlauthorlist}
\icmlauthor{Stefan Falkner}{ALUFR}
\icmlauthor{Aaron Klein}{ALUFR}
\icmlauthor{Frank Hutter}{ALUFR}
\end{icmlauthorlist}

\icmlaffiliation{ALUFR}{Department of Computer Science, University of Freiburg, Freiburg, Germany}

\icmlcorrespondingauthor{Stefan Falkner}{sfalkner@informatik.uni-freiburg.de}

\icmlkeywords{Machine Learning, ICML}

\vskip 0.3in
]



\printAffiliationsAndNotice{}  

\appendix

\section{Available Software}
To promote reproducible science and enable other researchers to use our method, we provide an open-source implementation of {\ourmethod} and Hyperband. It is available under \href{https://github.com/automl/HpBandSter}{https://github.com/automl/HpBandSter}.
The benchmarks and our scripts used to produce the data shown in the paper can be found in the \emph{icml\_2018} branch.

\section{Comparison to other Combinations of Bayesian optimization and Hyperband}

Here we discuss the differences between our method and the related approaches of \citet{bertrandhyperparameter} and \citet{wang-arxiv18} in more detail. We note that these works are independent and concurrent; our work extends our preliminary short workshop papers at NIPS 2017~\cite{falkner-bayesopt17} and ICLR 2018~\cite{falkner2018practical}.   

While the general idea of combining Hyperband and Bayesian optimization by \citet{bertrandhyperparameter} is the same as in our work, they use a Gaussian process for modeling the performance. The budget is modeled like any other dimension of the search space, without any special treatment. Based on our experience with Fabolas~\cite{klein-aistats17}, we expect that the squared exponential kernel might not extrapolate well, which would hinder performance. Also, the small evaluation provided by \citet{bertrandhyperparameter} does not allow strong conclusions about the performance of their method. 

\citet{wang-arxiv18} also independently combined TPE and Hyperband, but in a slightly different way than we did. In their method, TPE is used as a subroutine in every iteration of Hyperband. In particular, a new model is built from scratch at the beginning of every SuccessiveHalving run (Algorithm 3, line 8 in \citet{wang-arxiv18}). This means that in later iterations of the algorithm, the model does not benefit from any of the evaluations in previous iterations. In contrast, {\ourmethod} collects all evaluations on all budgets and uses the largest budget with enough evaluations (admittedly a heuristic, but we would argue a reasonable one) as a base for future evaluations. This way, {\ourmethod} aggregates more knowledge into its models for the different budgets as the optimization progresses. We believe this to be a crucial part of the strong performance of our method. Empirically, \citet{wang-arxiv18} did not achieve the consistent and large speedups across a wide range of applications \ourmethod{} achieved in our experiments.

\section{Successive Halving}
SuccessiveHalving is a simple heuristic to allocate more resources to promising candidates.
For completeness, we provide pseudo code for it in Algorithm \ref{alg:SH}.
It is initialized with a set of configurations, a minimum and maximum budget, and a scaling parameter $\eta$.
In the first \emph{stage} all configurations are evaluated on the smallest budget (line 3).
The losses are then sorted and only the best $1/\eta$ configurations are kept in the set $C$ (line 4).
For the following stage, the budget is increased by a factor of $\eta$ (line 5).
This is repeated until the maximum budget for a single configuration is reached (line 2).
Within Hyperband, the budgets are chosen such that all SuccessiveHalving executions require a similar total budget.

\begin{algorithm2e}[ht]
	\caption{Pseudocode for SuccessiveHalving used by Hyperband as a subroutine.}\label{alg:SH}
        \DontPrintSemicolon
        \SetKwInOut{Input}{input}\SetKwInOut{Output}{output}
	\Input{initial budget $b_0$, maximum budget $b_{max}$, set of $n$ configurations $C = \{ c_1, c_2,\dots c_n\}$}
	$b = b_0$\;
	\While{$ b \leq b_{max}$}{
		$L = \{\tilde{f}(c, b) : c \in C\}$\;
		$C = \mathrm{top_k}(C, L, \lfloor \vert C\vert / \eta) \rfloor $\;
		$b = \eta \cdot b$
	}
\end{algorithm2e}

\begin{figure*}[t!]
	\centering
	\includegraphics[width=0.45\linewidth]{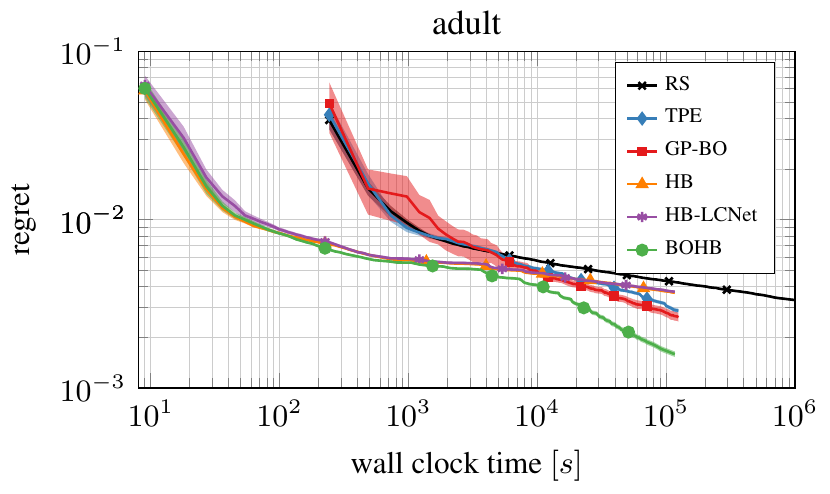}
	\includegraphics[width=0.45\linewidth]{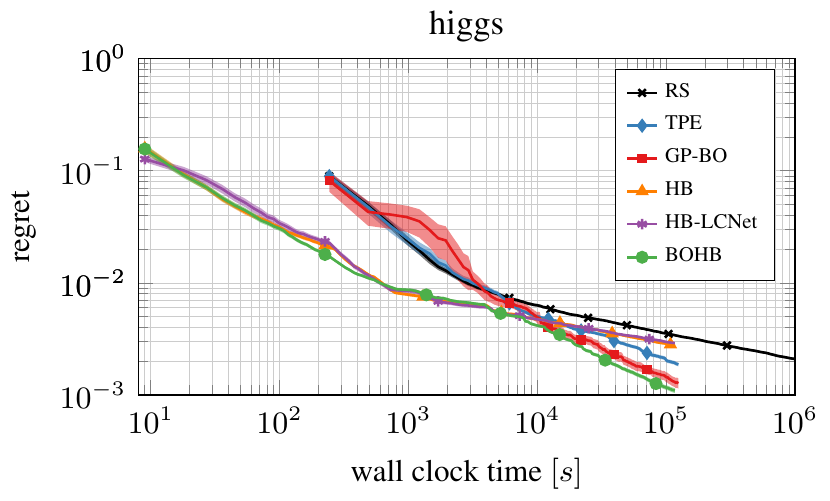}
	\includegraphics[width=0.45\linewidth]{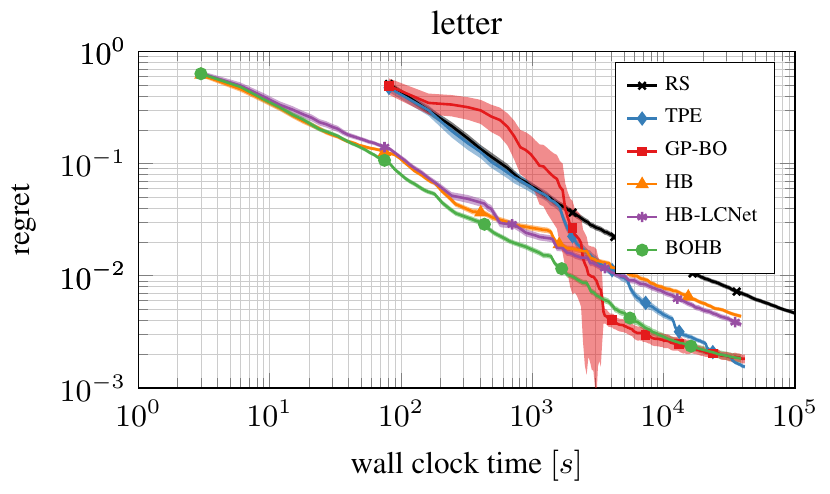}
	\includegraphics[width=0.45\linewidth]{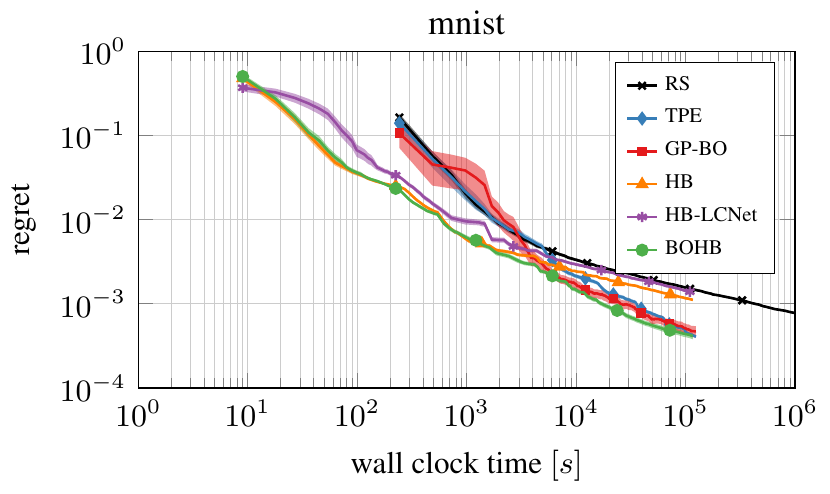}
	\includegraphics[width=0.45\linewidth]{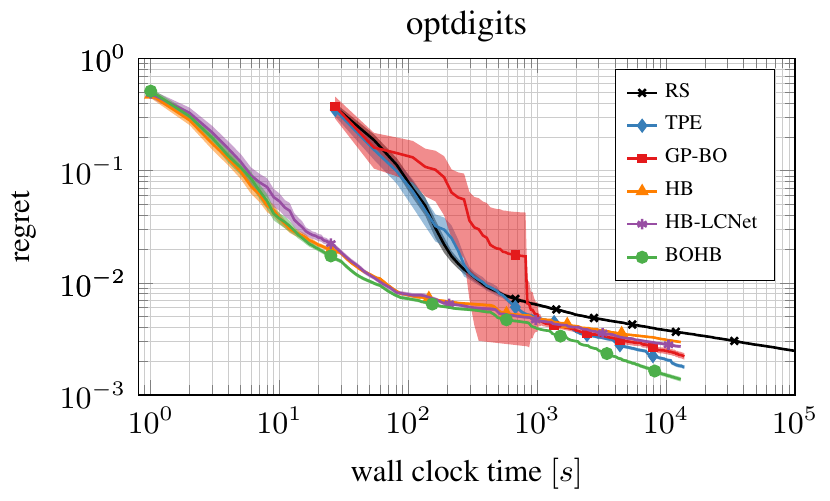}
	\includegraphics[width=0.45\linewidth]{pgf_plots/surrogates/methods_poker.pdf}
	\caption{Mean performance on the surrogates for all six datasets. As uncertainties, we show the standard error of the mean based on 512 runs (except for GP-BO, which has only 50 runs).}
	\label{fig:paramnet_methods_all}
\end{figure*}

\section{Details on the Kernel Density Estimator}
We used the MultivariateKDE from the statsmodels package \cite{seabold2010statsmodels}, which constructs a factorized kernel, with a one-dimensional kernel for each dimension.
Note that using this product of 1-d kernels differs from the original TPE, which uses a pdf that is the product of 1-d pdfs.
For the continuous parameters a Gaussian kernel is used, whereas the Aitchison-Aitken kernel is the default for categorical parameters. 
We used Scott’s rule for efficient bandwidth estimation, as preliminary experiments with maximum-likelihood based bandwidth selection did not yield better performance but caused a significant overhead.

\section{Performance of all methods on all surrogates}

Figure \ref{fig:paramnet_methods_all} shows the performance of all methods we evaluated on all our surrogate benchmarks.
Random search is clearly the worst optimizer across all datasets when the budget is large enough for GP-BO and TPE to leverage their model.
Hyperband and the two methods based on it (HB-LCNet) and {\ourmethod} improve much more quickly due to the smaller budgets used.
On all surrogate benchmarks, {\ourmethod} starts to outperform HB after the first couple of iterations (sometimes even earlier, e.g., on dataset letter).
The same dataset also shows that traditional BO methods can still have an advantage for very large budgets, since in these late stages of the optimization process the low fidelity evaluations of {\ourmethod} can cause a constant overhead without any gain.

\begin{figure*}[ht]
	\centering
	\includegraphics[width=0.45\linewidth]{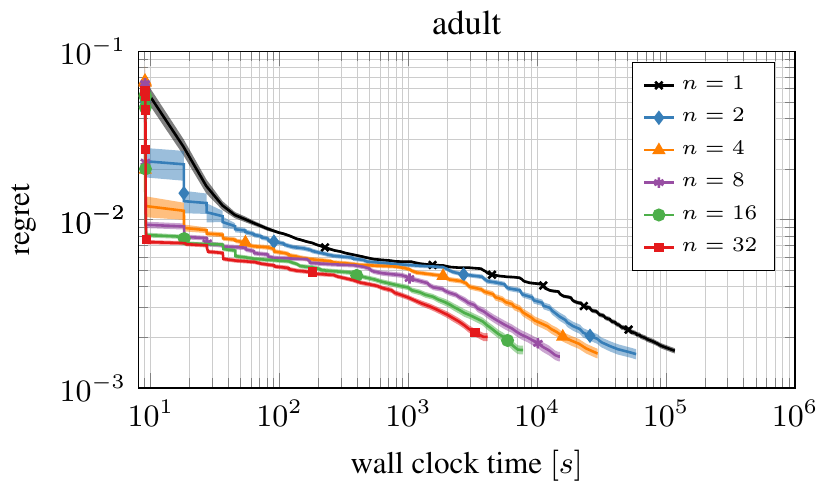}
	\includegraphics[width=0.45\linewidth]{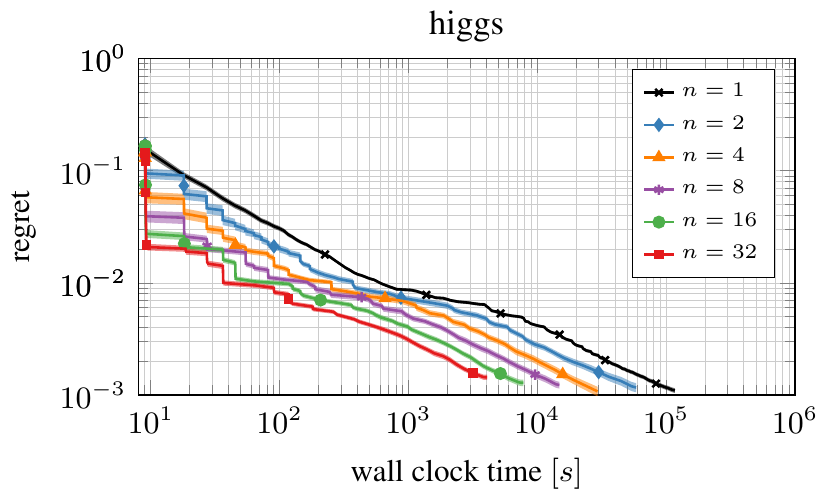}
	\includegraphics[width=0.45\linewidth]{pgf_plots/surrogates/parallelization_letter.pdf}
	\includegraphics[width=0.45\linewidth]{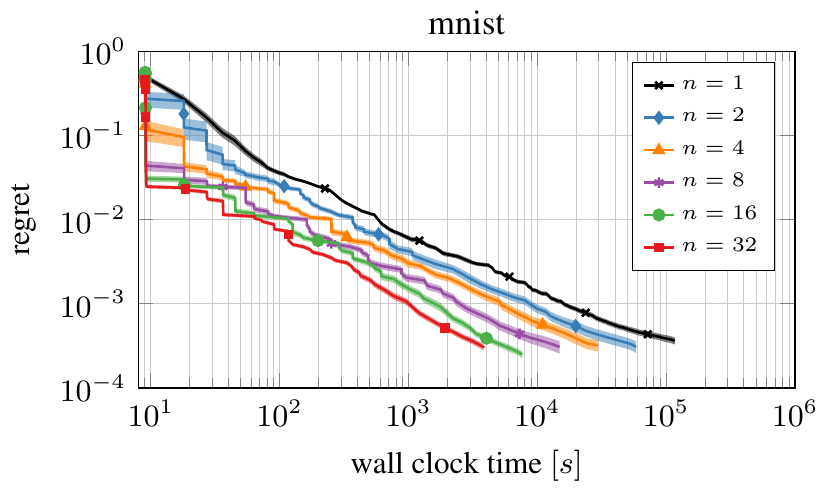}
	\includegraphics[width=0.45\linewidth]{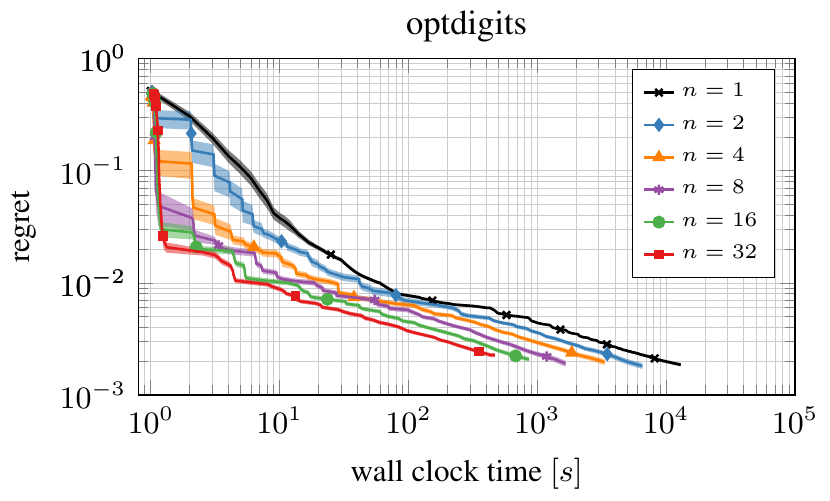}
	\includegraphics[width=0.45\linewidth]{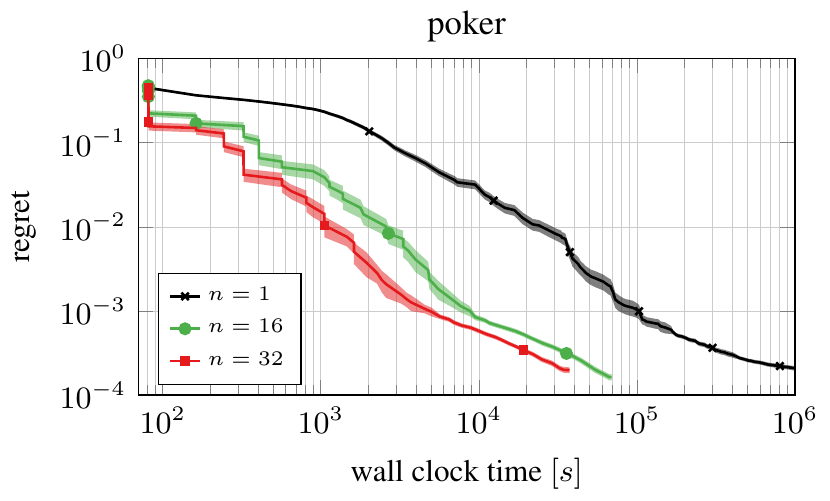}
	\caption{Mean performance on the surrogates for all six datasets with different numbers of workers $n$. As uncertainties, we show the standard error of the mean based on 128 runs. Because we simulated them in real time to capture the true behavior, poker is too expensive to evaluate with less than 16 workers within a day.}
	\label{fig:paramnet_parallel_all}
\end{figure*}

\section{Performance of parallel runs}
Figure \ref{fig:paramnet_parallel_all} shows the performance of {\ourmethod} when run in parallel on all our surrogate benchmarks.
The speed-ups are quite consistent, and almost linear for a small number of workers (2-8).
For more workers, more random configurations are evaluated in parallel before the first model is built, which degrades performance. But even for 32 workers, linear speedups are possible (see, e.g., dataset letter, for reaching a regret of $2 \times 10^{-3}$).

We note that in order to carry out this evaluation of parallel performance, we actually simulated the parallel optimization by making each worker wait for the given budget before returning the corresponding performance value of our surrogate benchmark.
(The case of one worker is an exception, where we can simply reconstruct the trajectory because all configurations are evaluated serially.)
By using this approach in connection with threads, each evaluation of a parallel algorithm still only used 1 CPU, but the run actually ran in real time.
For this reason, we decided to not evaluate all possible numbers of workers for dataset poker, for which each run with less than 16 workers would have taken more than a day, and we do not expect any different behavior compared to the other datasets.

\begin{figure*}[t!]
	\centering
	\includegraphics[width=0.45\linewidth]{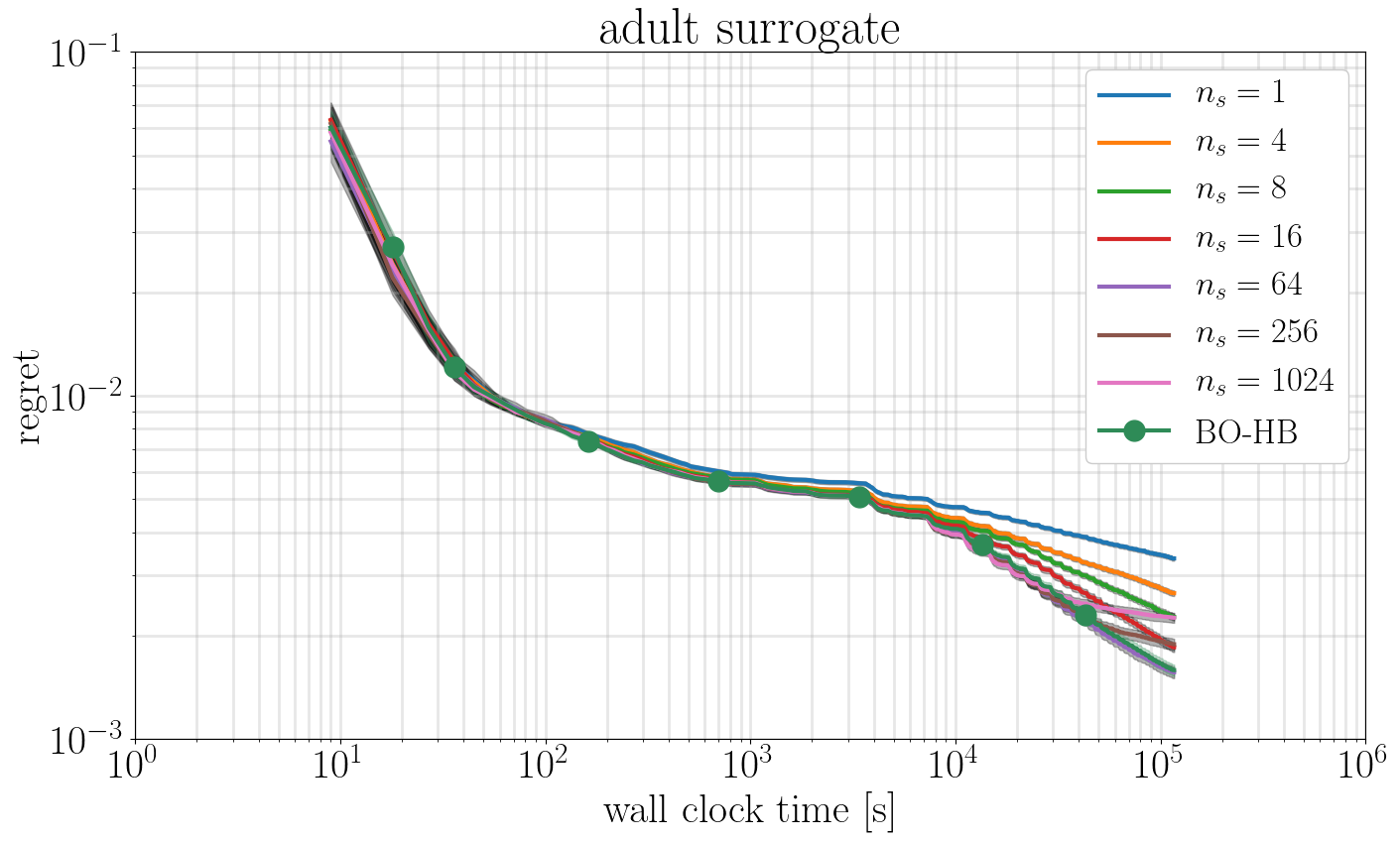}
	\includegraphics[width=0.45\linewidth]{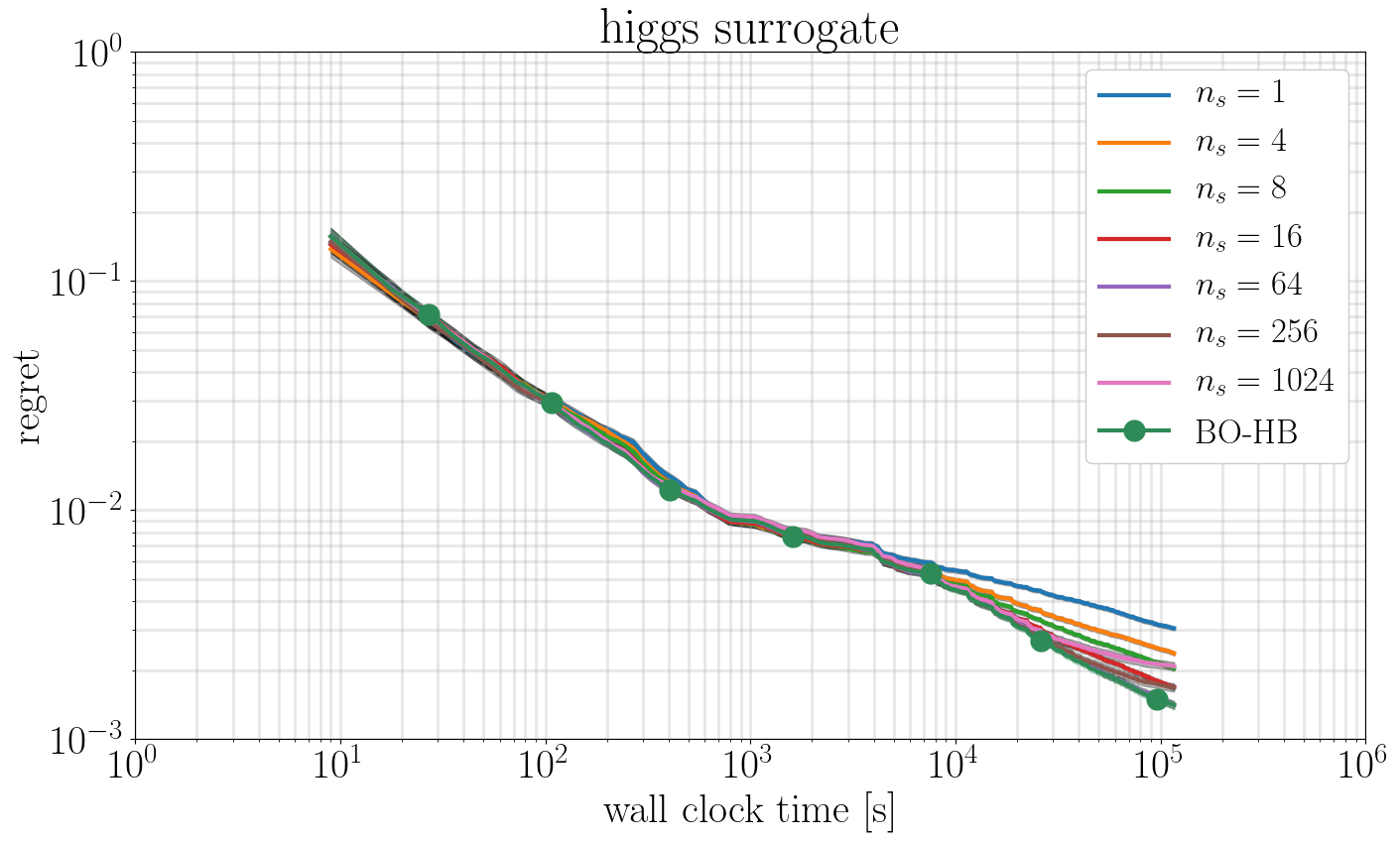}
	\includegraphics[width=0.45\linewidth]{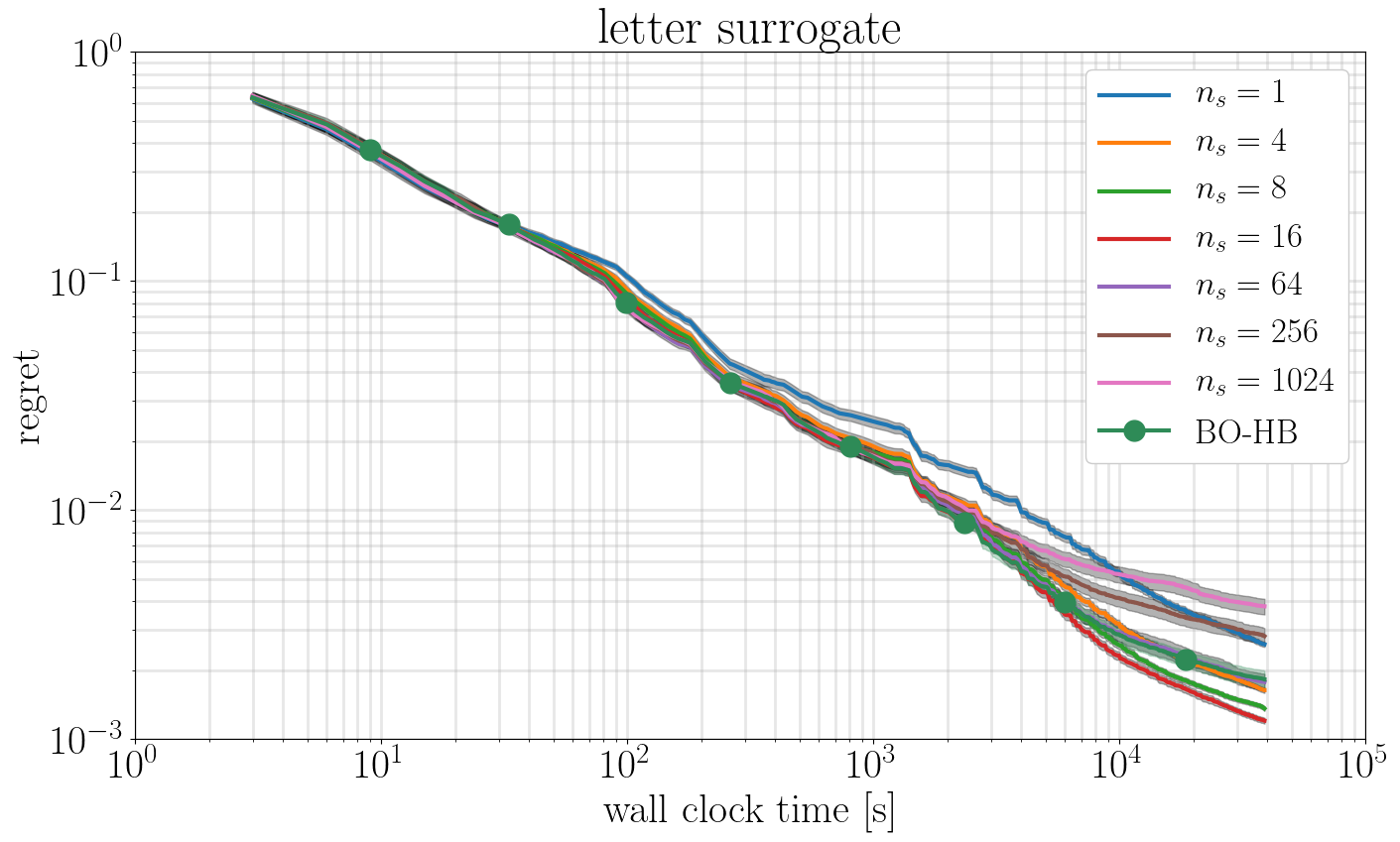}
	\includegraphics[width=0.45\linewidth]{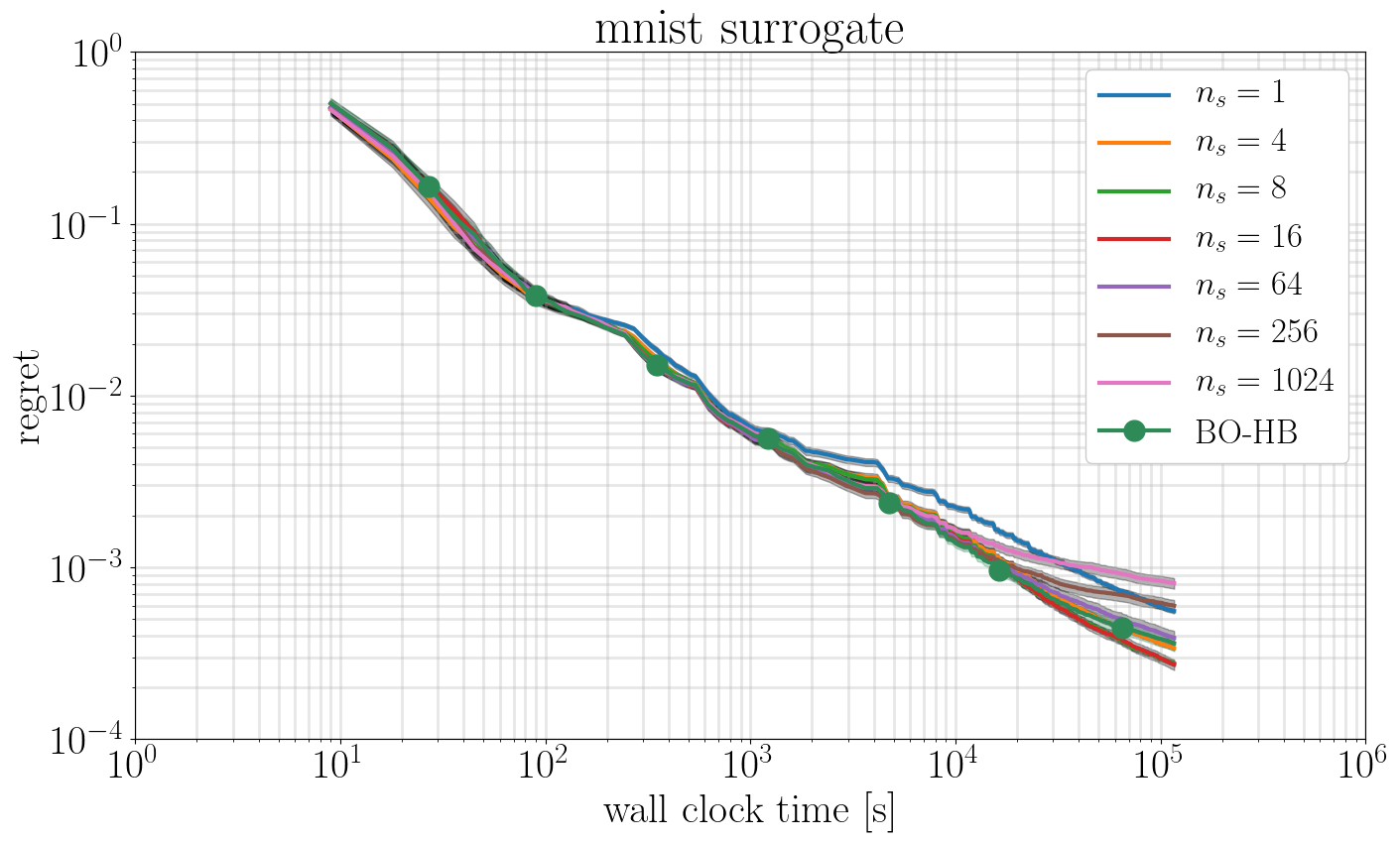}
	\includegraphics[width=0.45\linewidth]{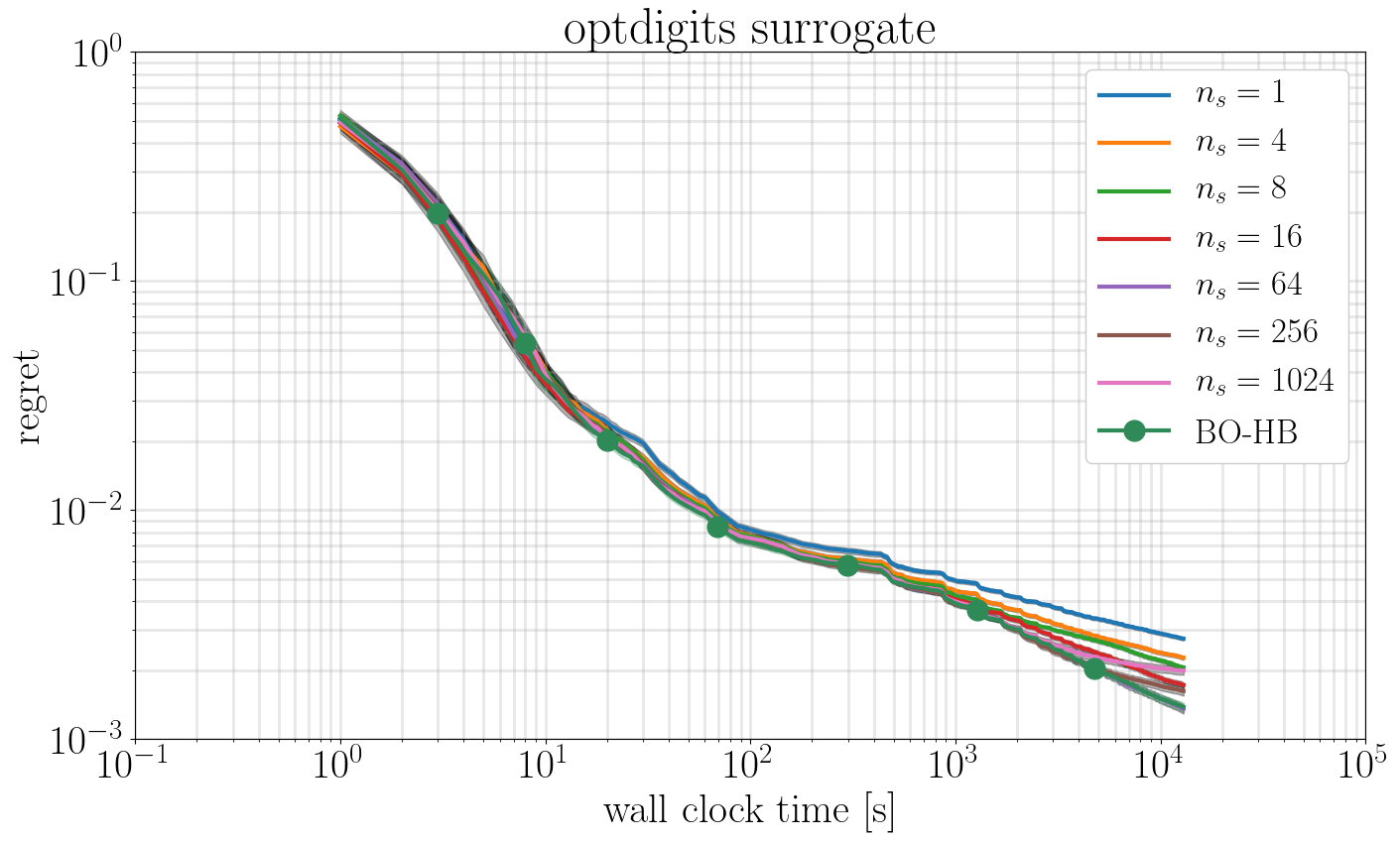}
	\includegraphics[width=0.45\linewidth]{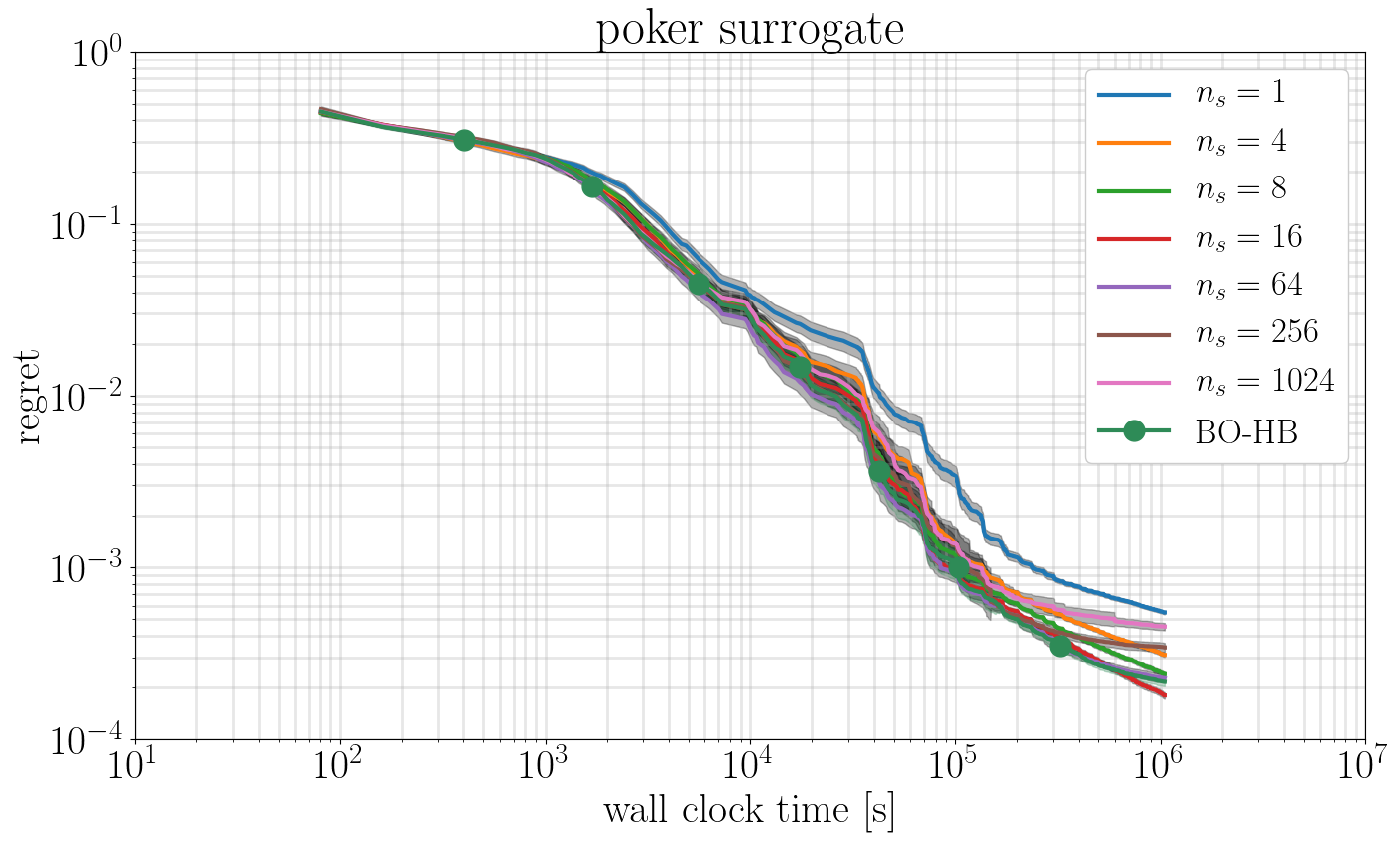}
	\caption{Performance on the surrogates for all six datasets for different number of samples}
	\label{fig:paramnet_num_samples_all}
\end{figure*}

\section{Evaluating the hyperparameters of {\ourmethod}}

\begin{figure*}[t]
	\centering
	\includegraphics[width=0.45\linewidth]{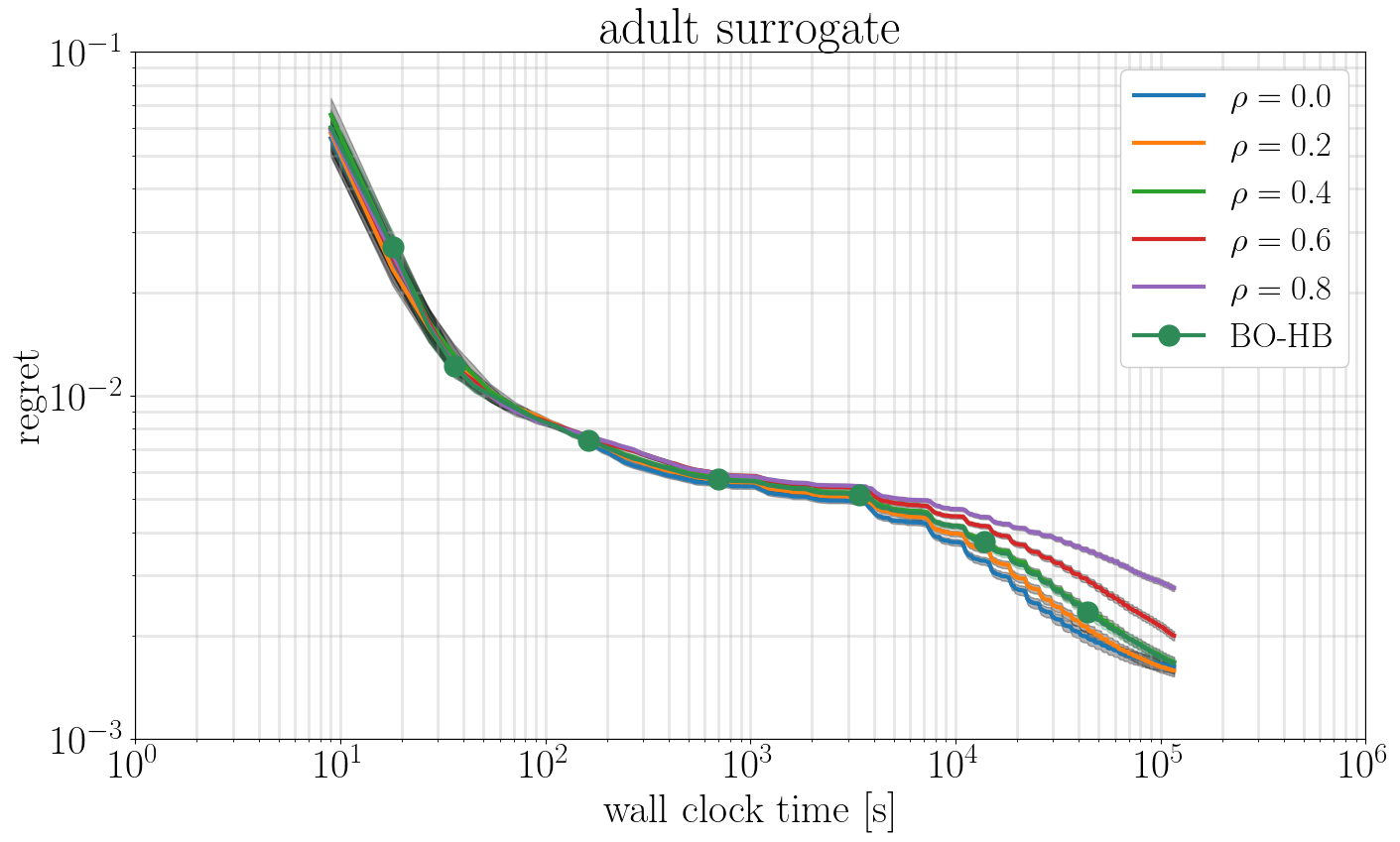}
	\includegraphics[width=0.45\linewidth]{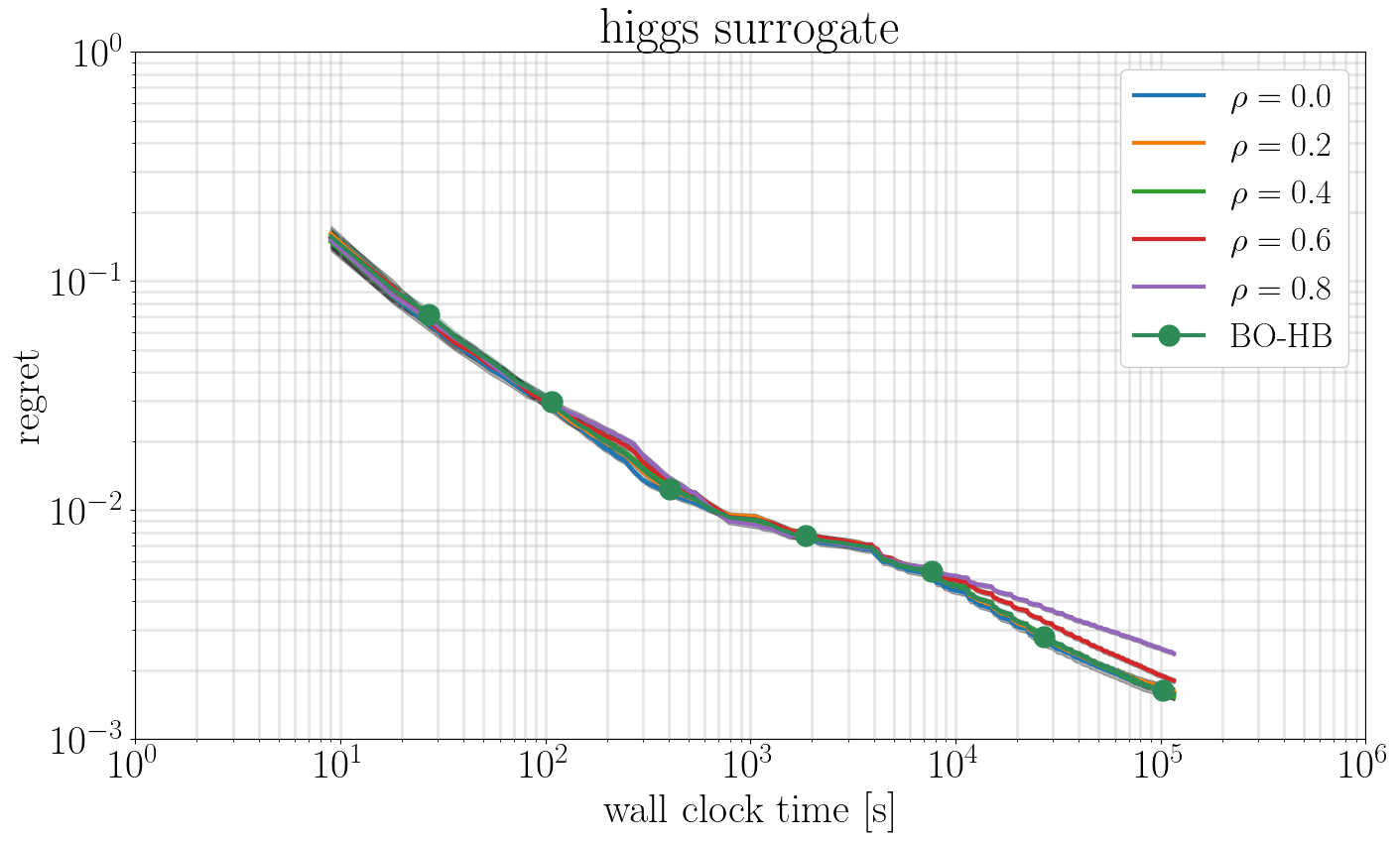}
	\includegraphics[width=0.45\linewidth]{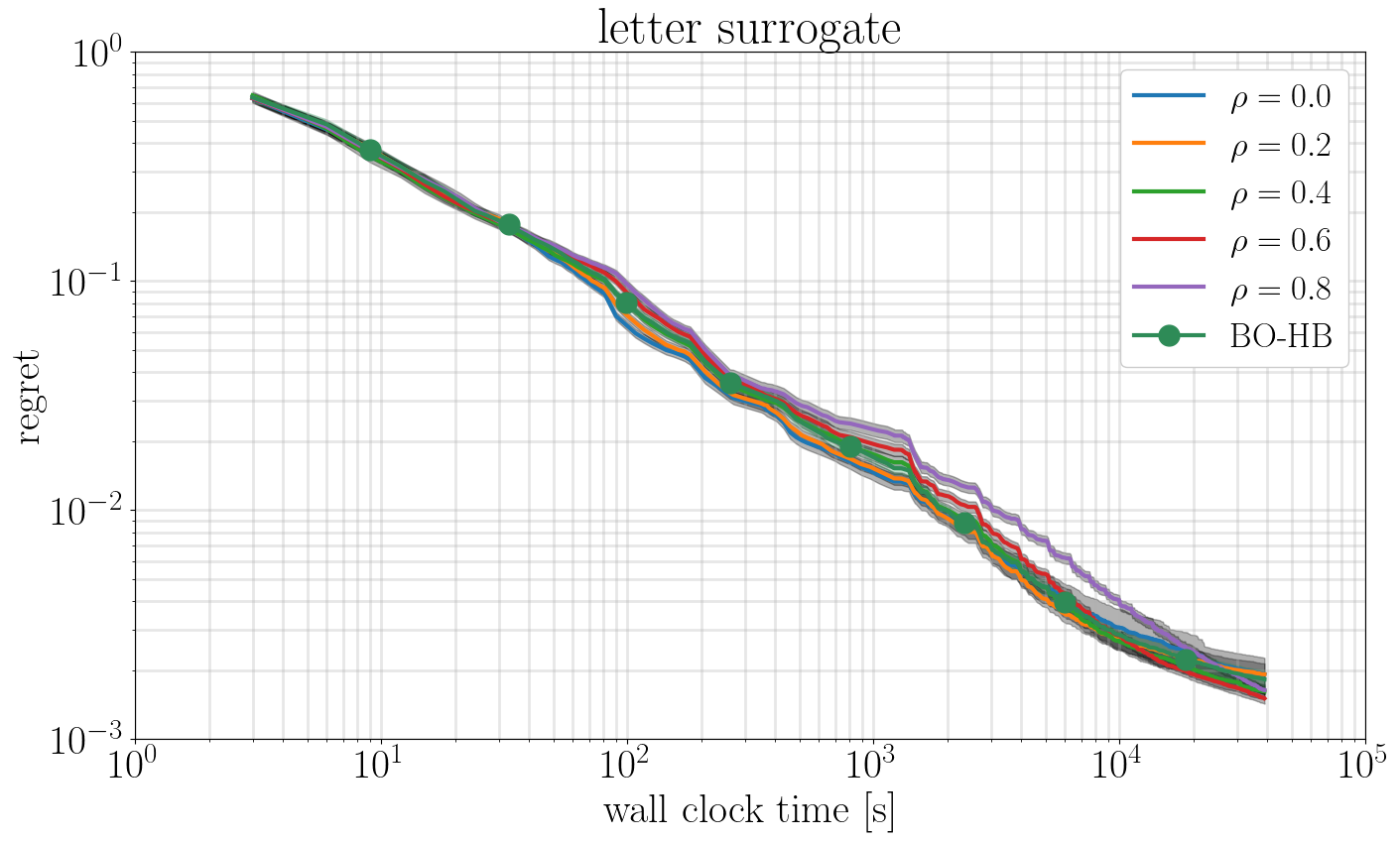}
	\includegraphics[width=0.45\linewidth]{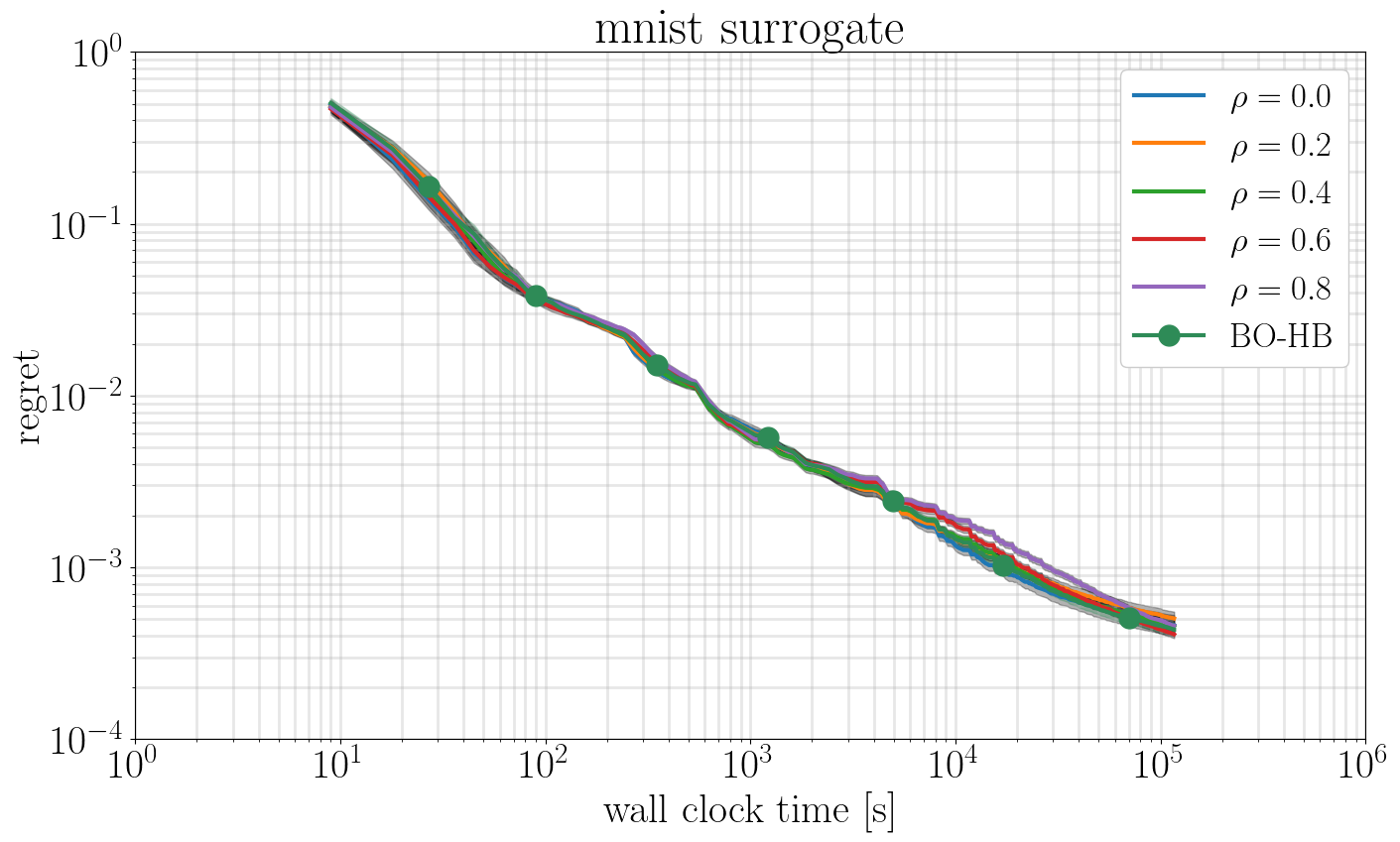}
	\includegraphics[width=0.45\linewidth]{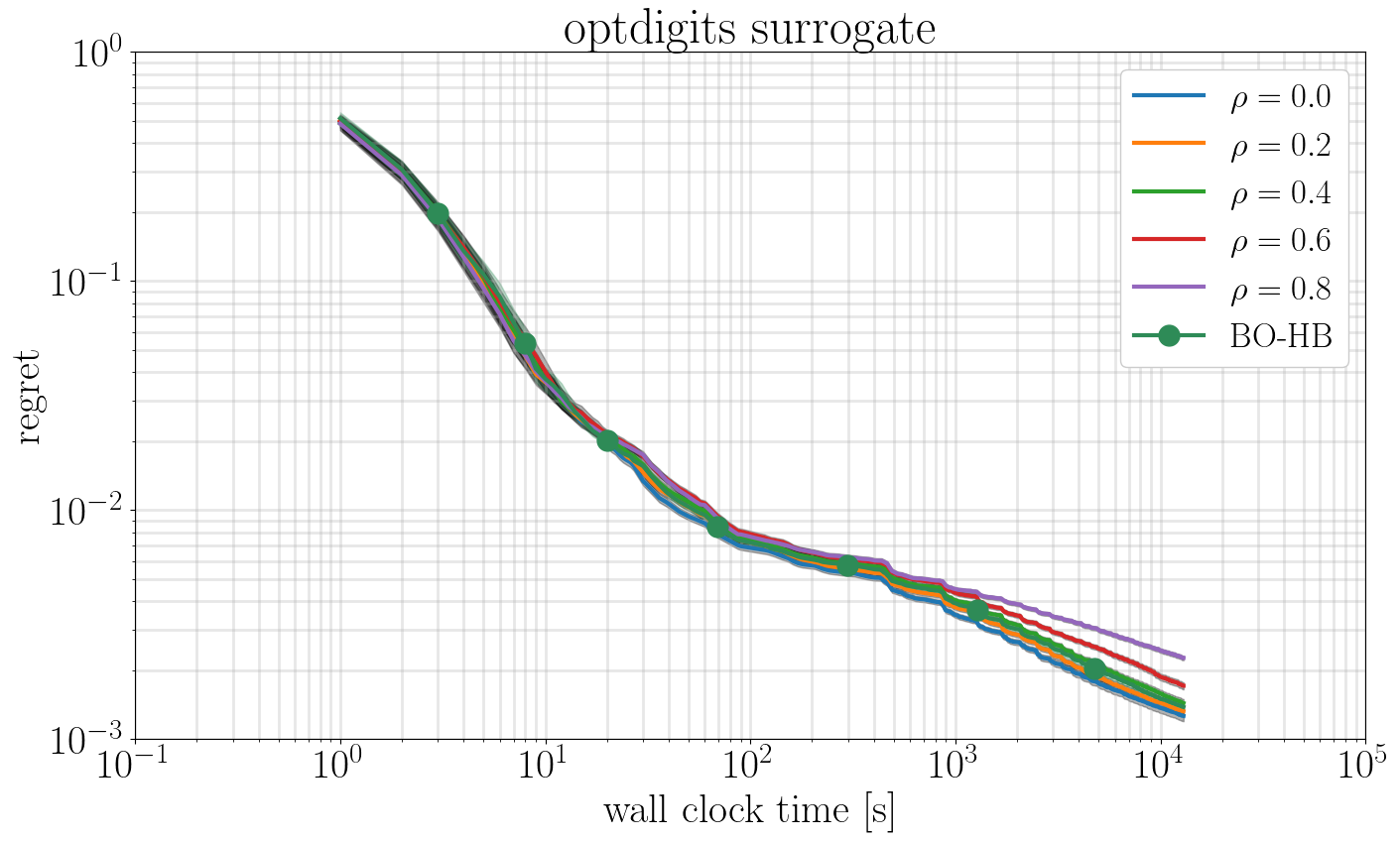}
	\includegraphics[width=0.45\linewidth]{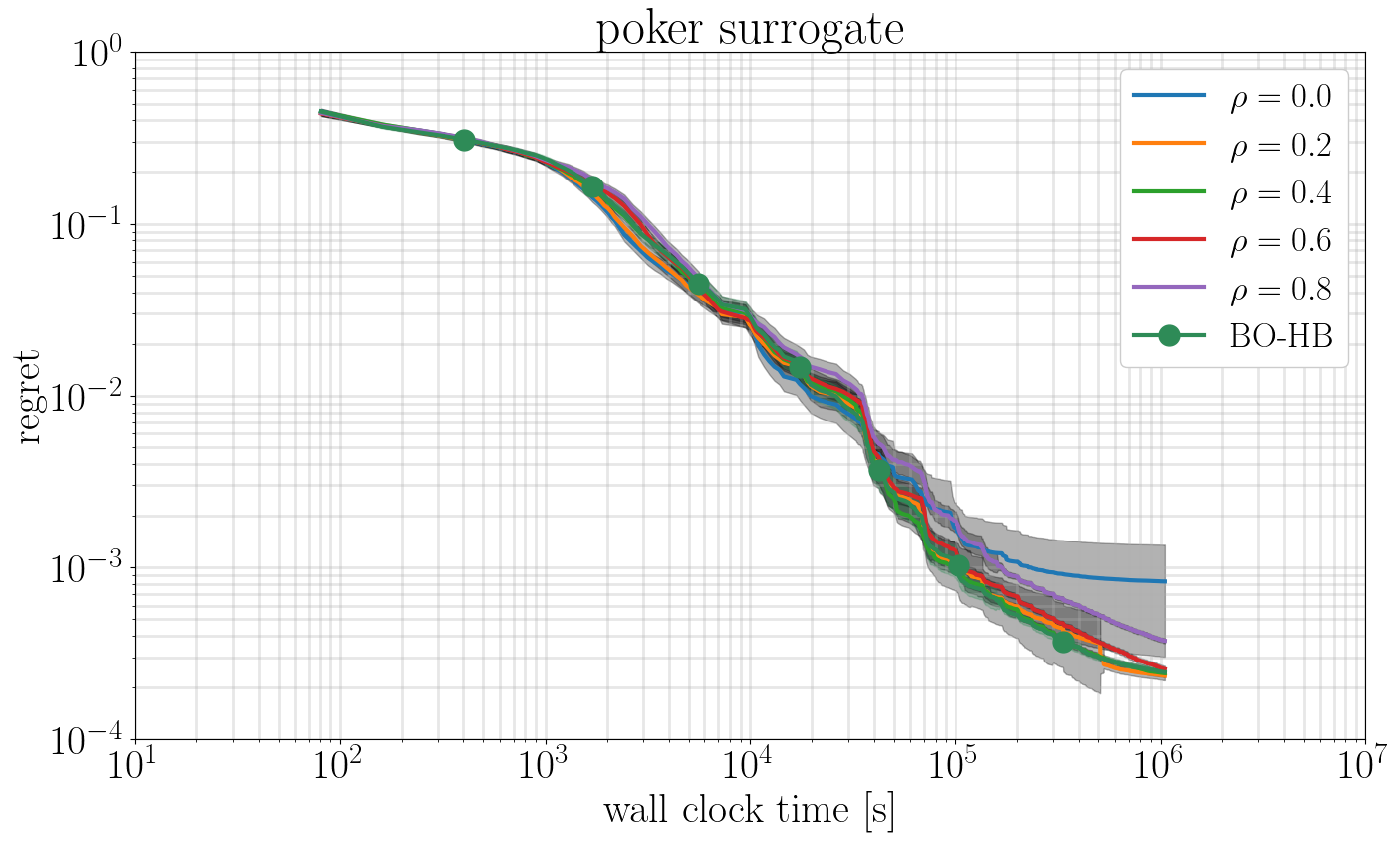}
	\caption{Performance on the surrogates for all six datasets for different random fractions}
	\label{fig:paramnet_random_fraction_all}
\end{figure*}

\begin{figure*}[t]
	\centering
	\includegraphics[width=0.45\linewidth]{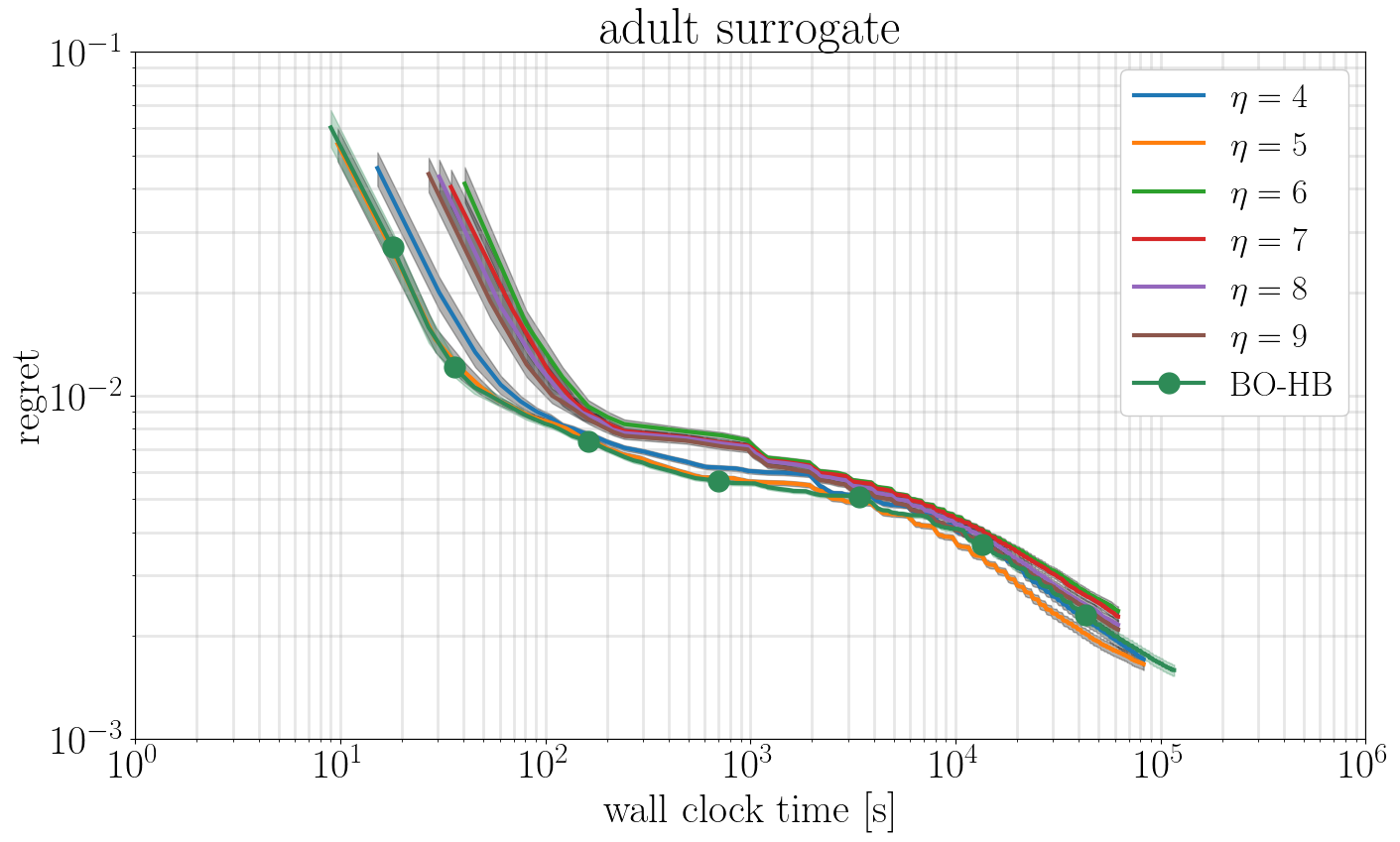}
	\includegraphics[width=0.45\linewidth]{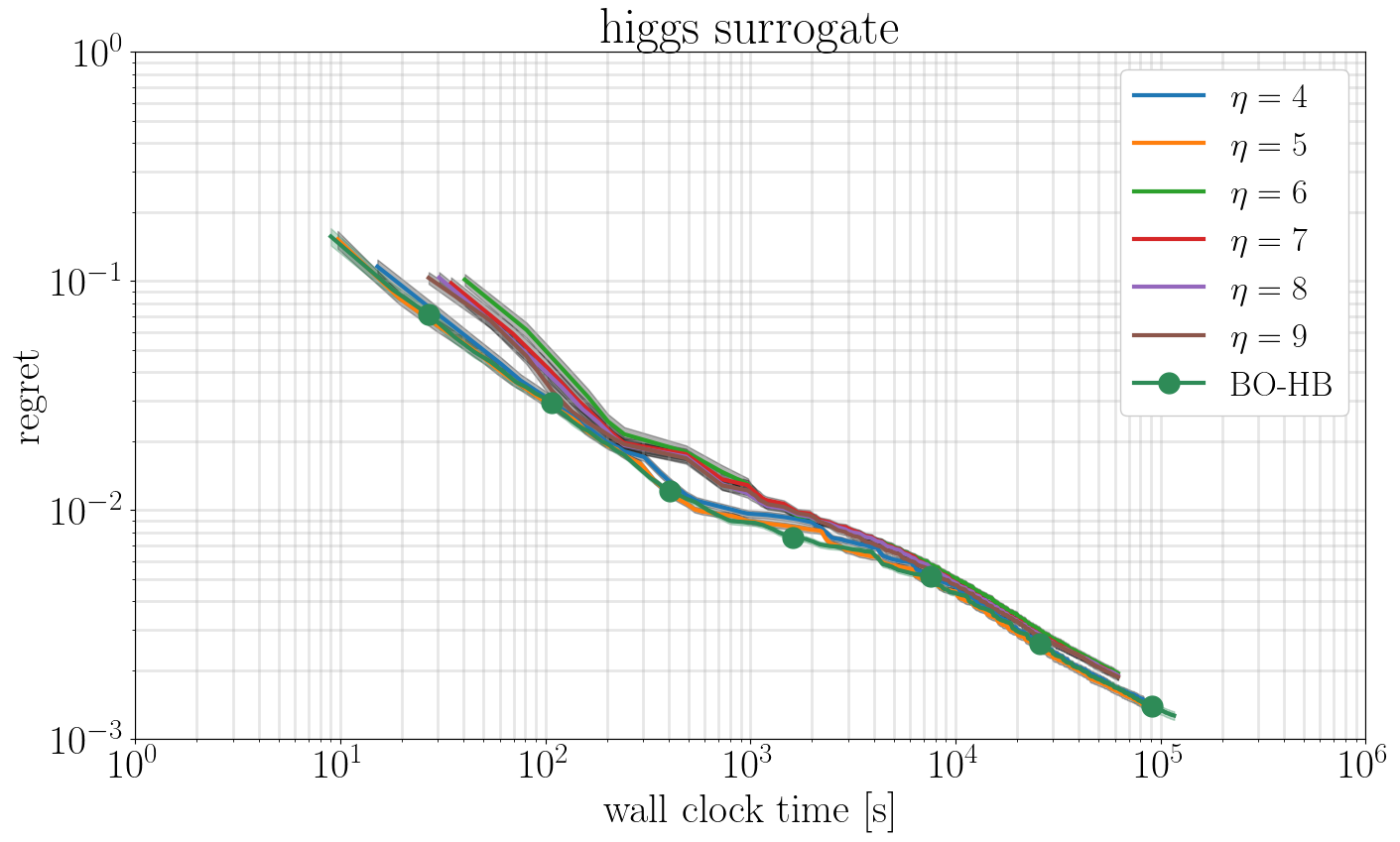}
	\includegraphics[width=0.45\linewidth]{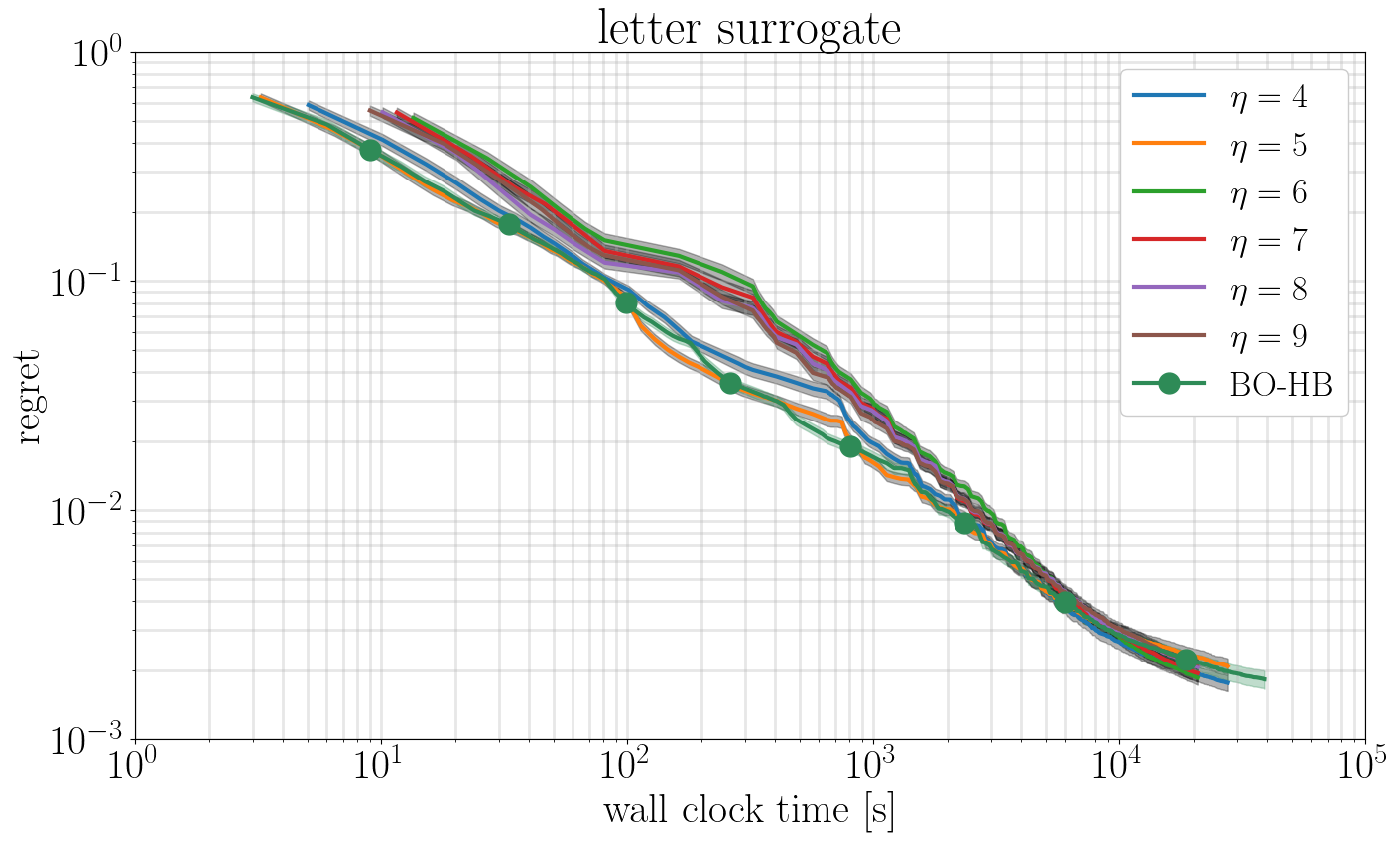}
	\includegraphics[width=0.45\linewidth]{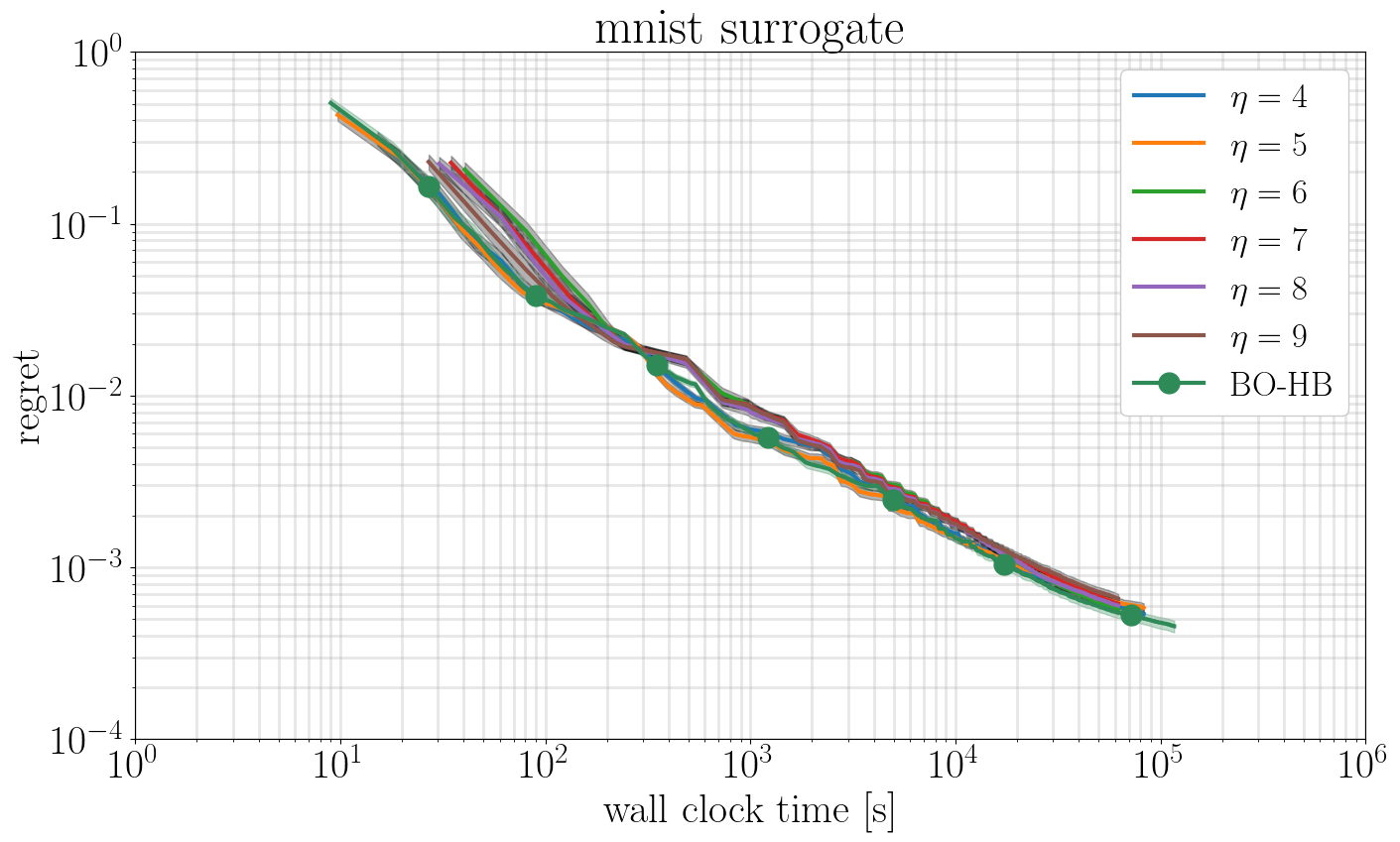}
	\includegraphics[width=0.45\linewidth]{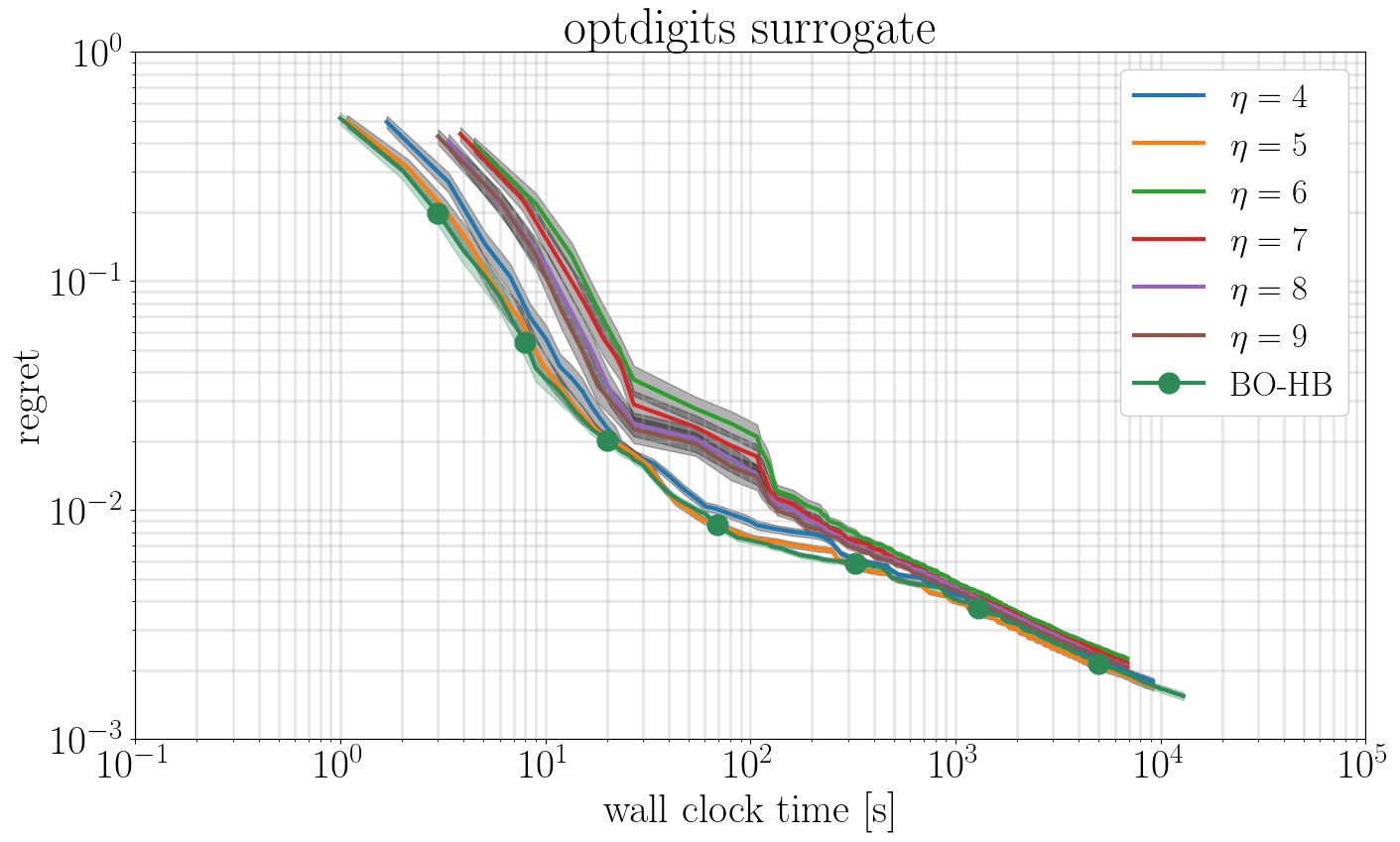}
	\includegraphics[width=0.45\linewidth]{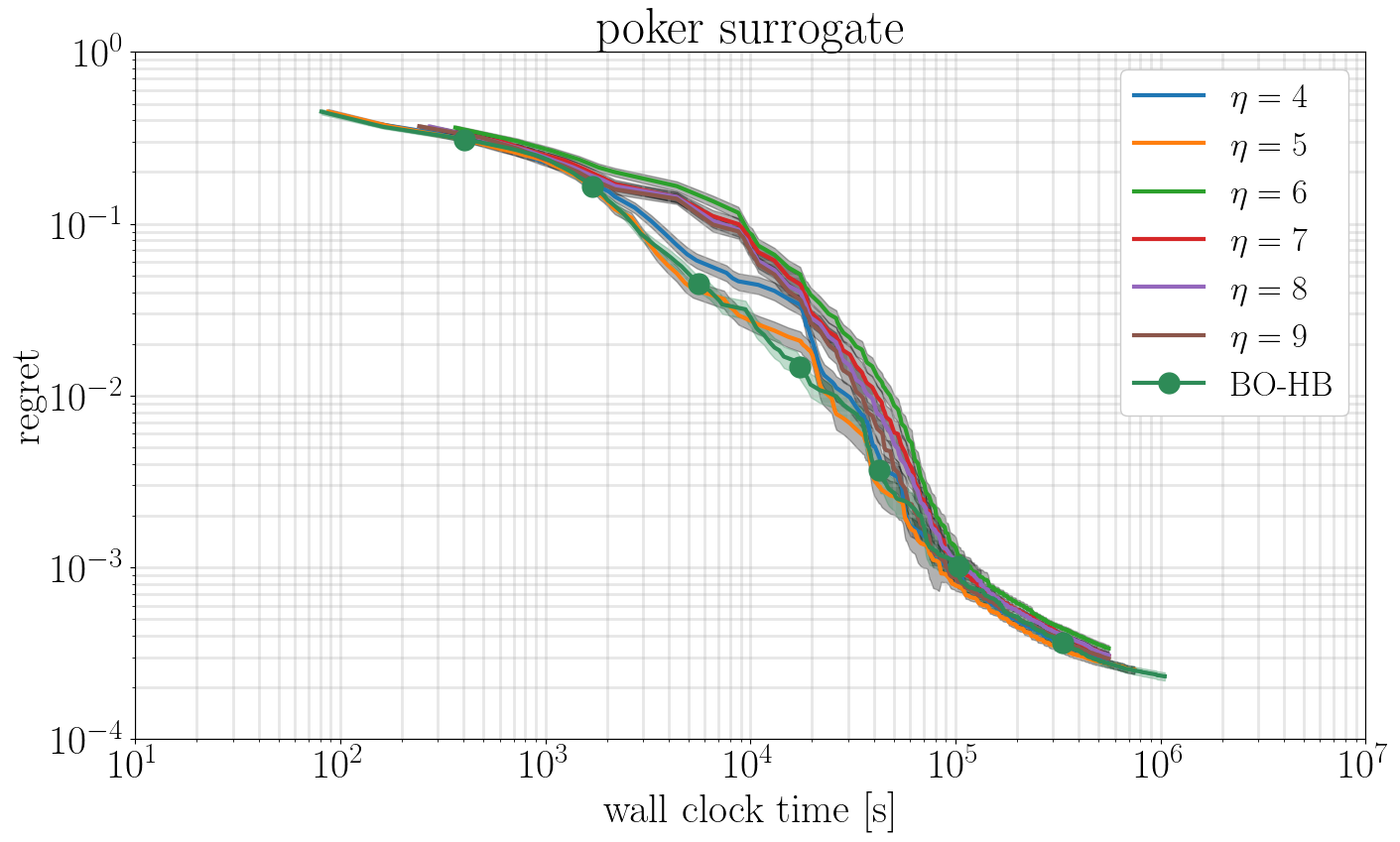}
	\caption{Performance on the surrogates for all six datasets for different values of $\eta$.}
	\label{fig:paramnet_etas_all}
\end{figure*}

\begin{figure*}[t]
	\centering
	\includegraphics[width=0.45\linewidth]{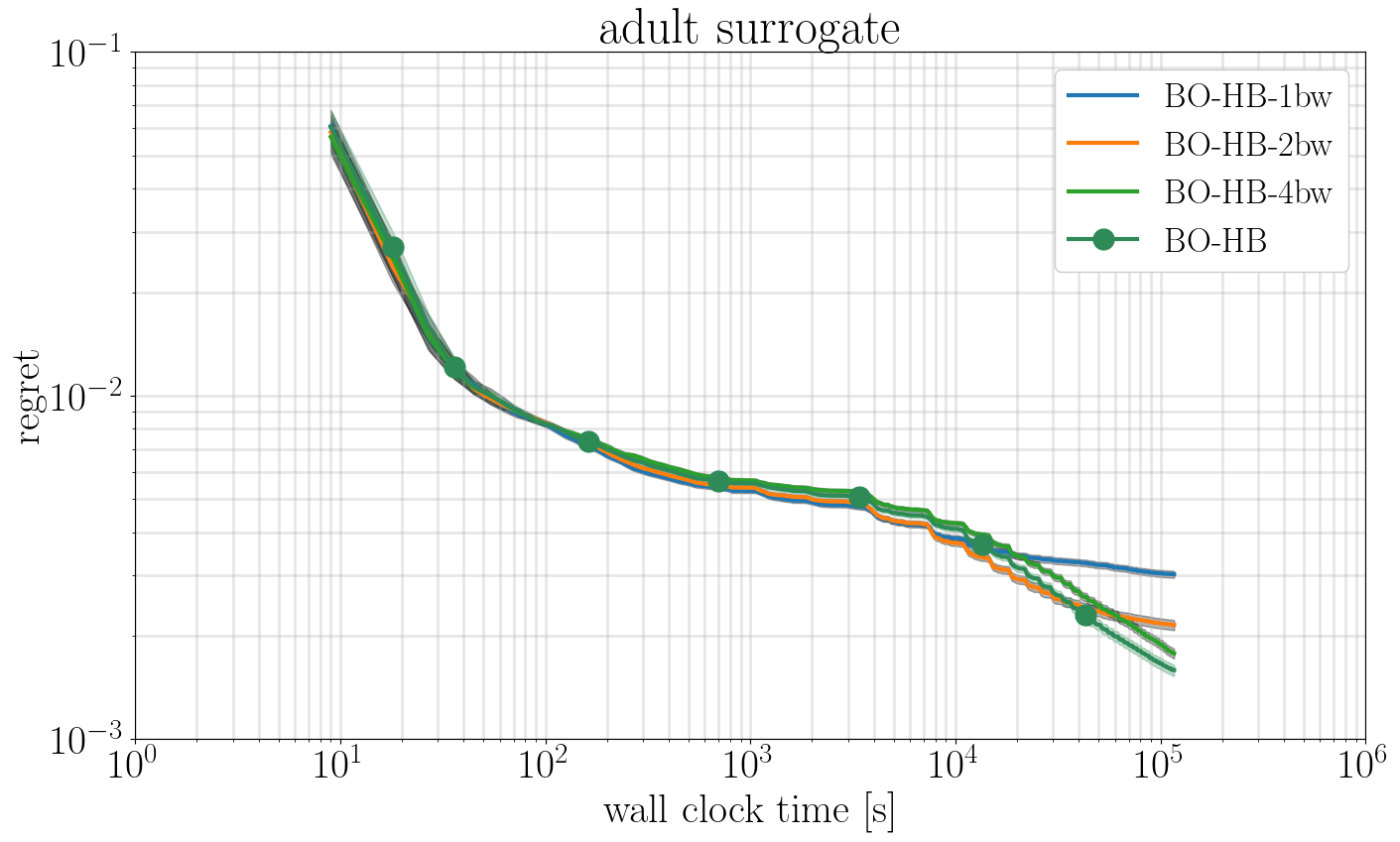}
	\includegraphics[width=0.45\linewidth]{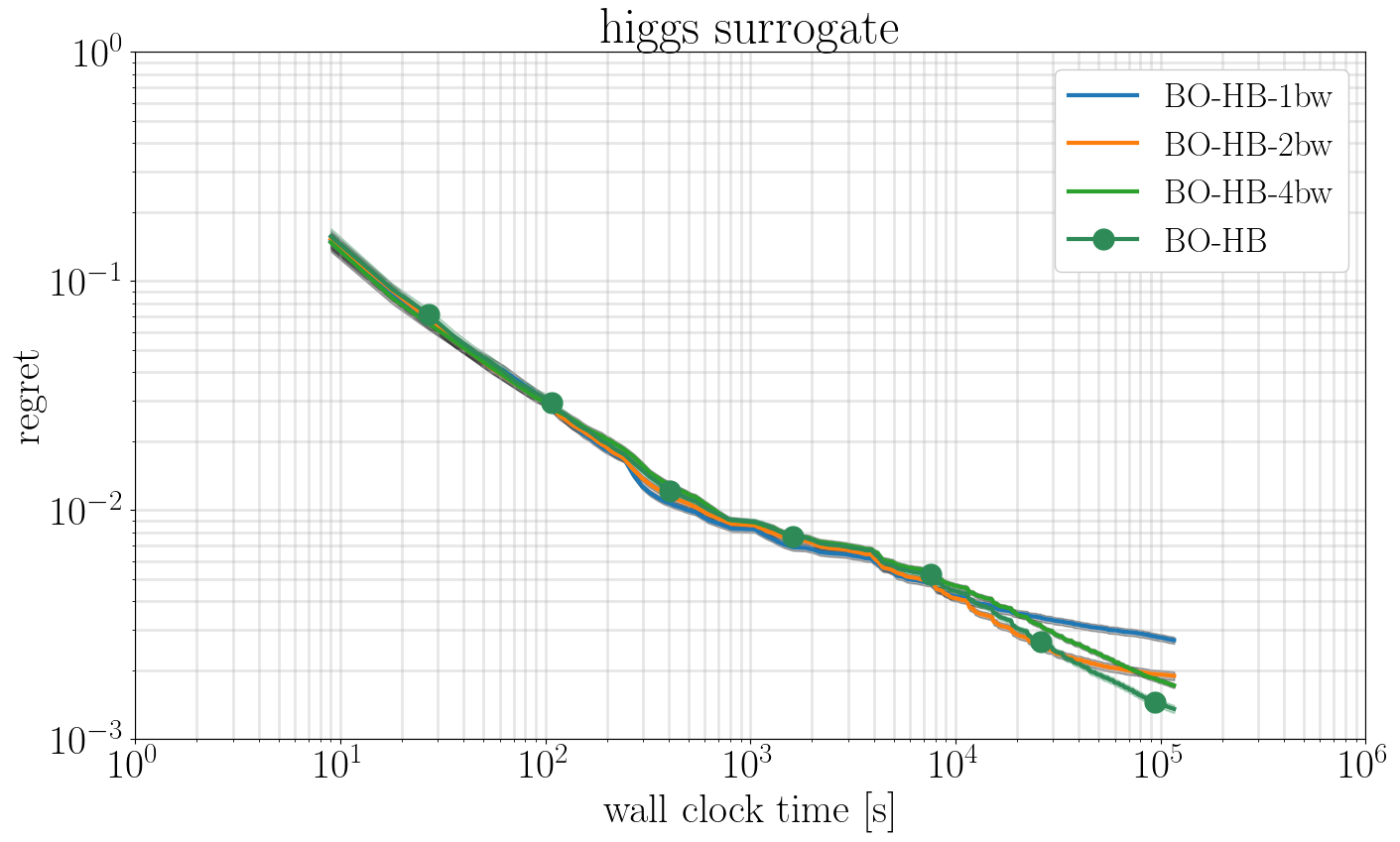}
	\includegraphics[width=0.45\linewidth]{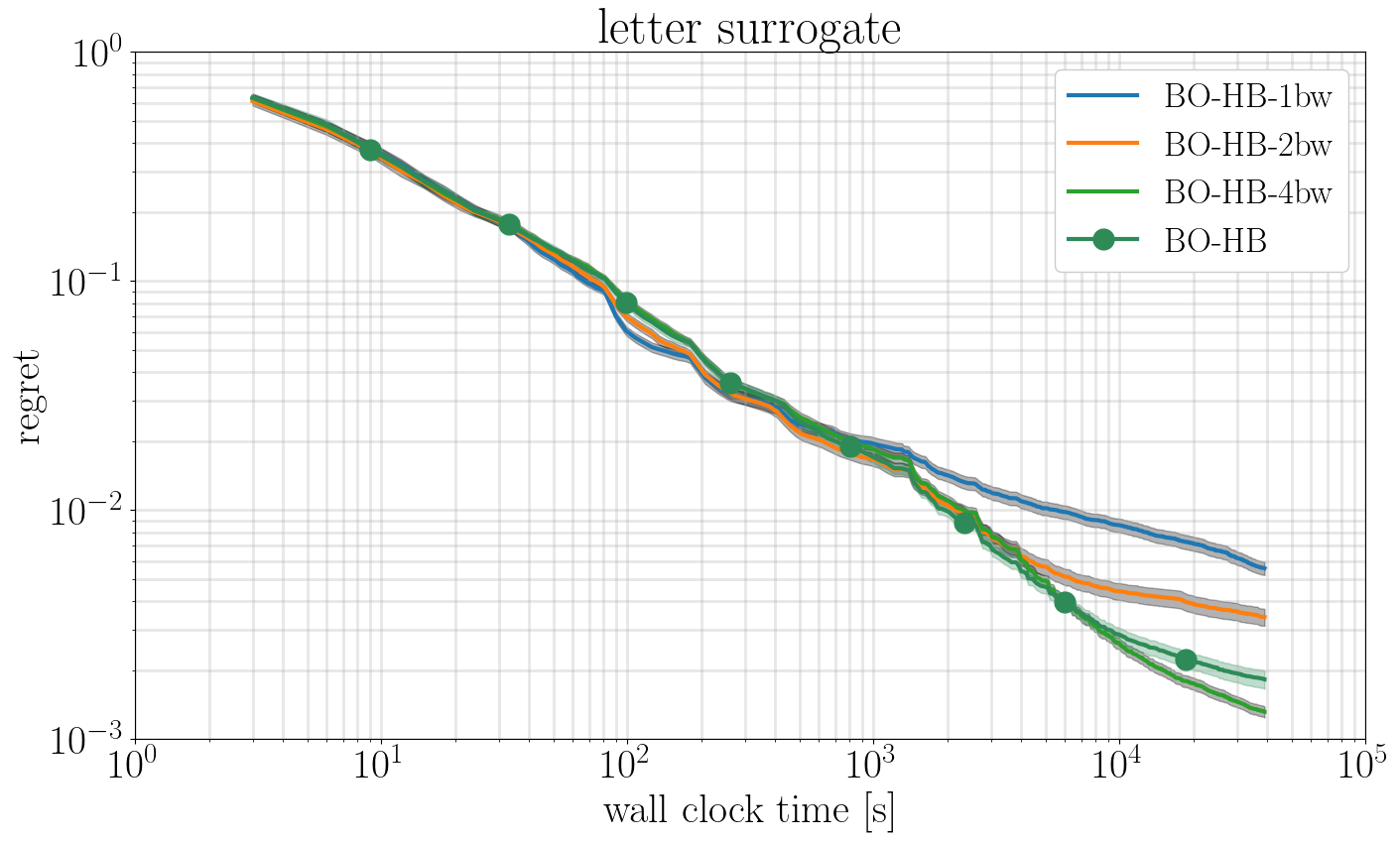}
	\includegraphics[width=0.45\linewidth]{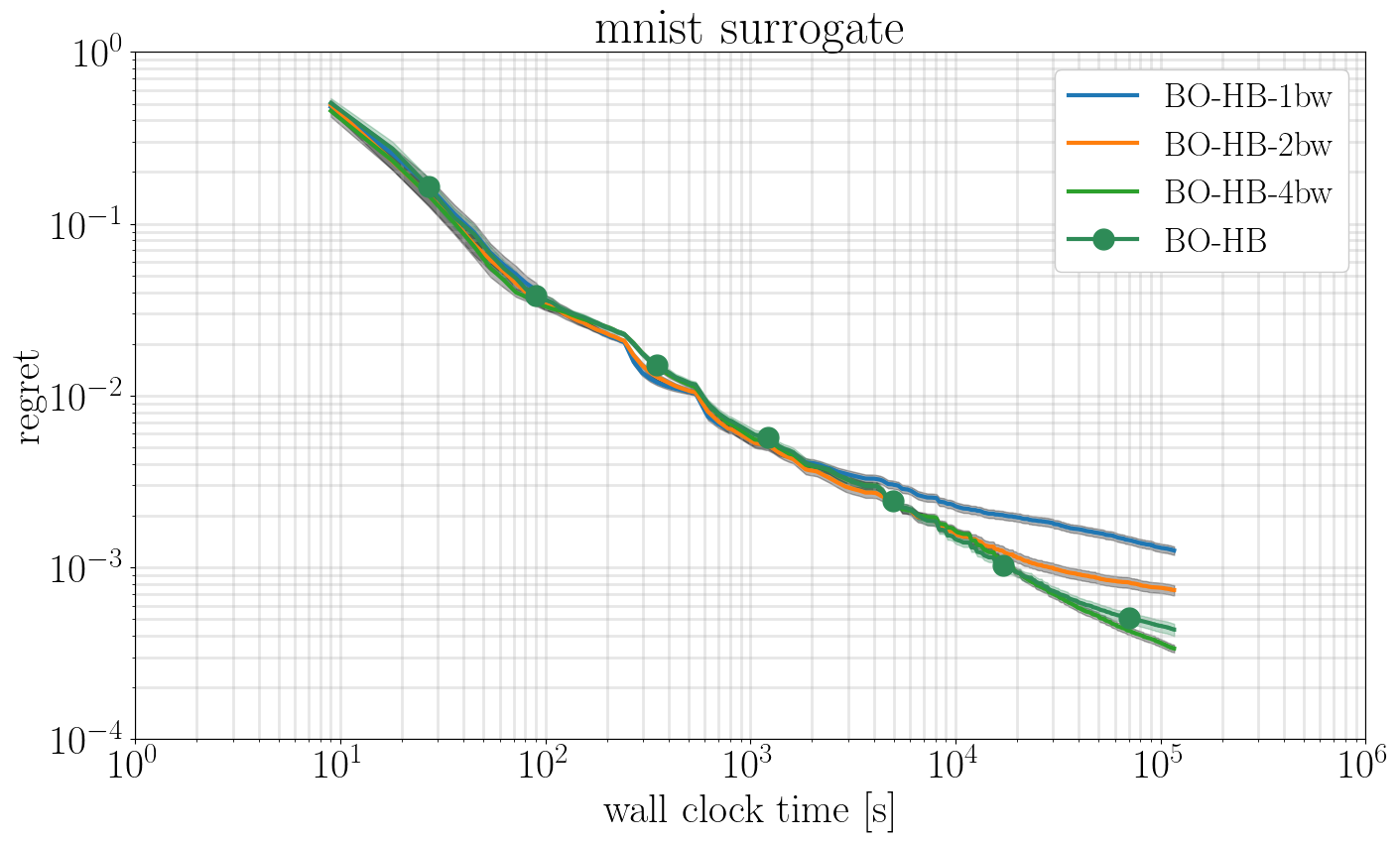}
	\includegraphics[width=0.45\linewidth]{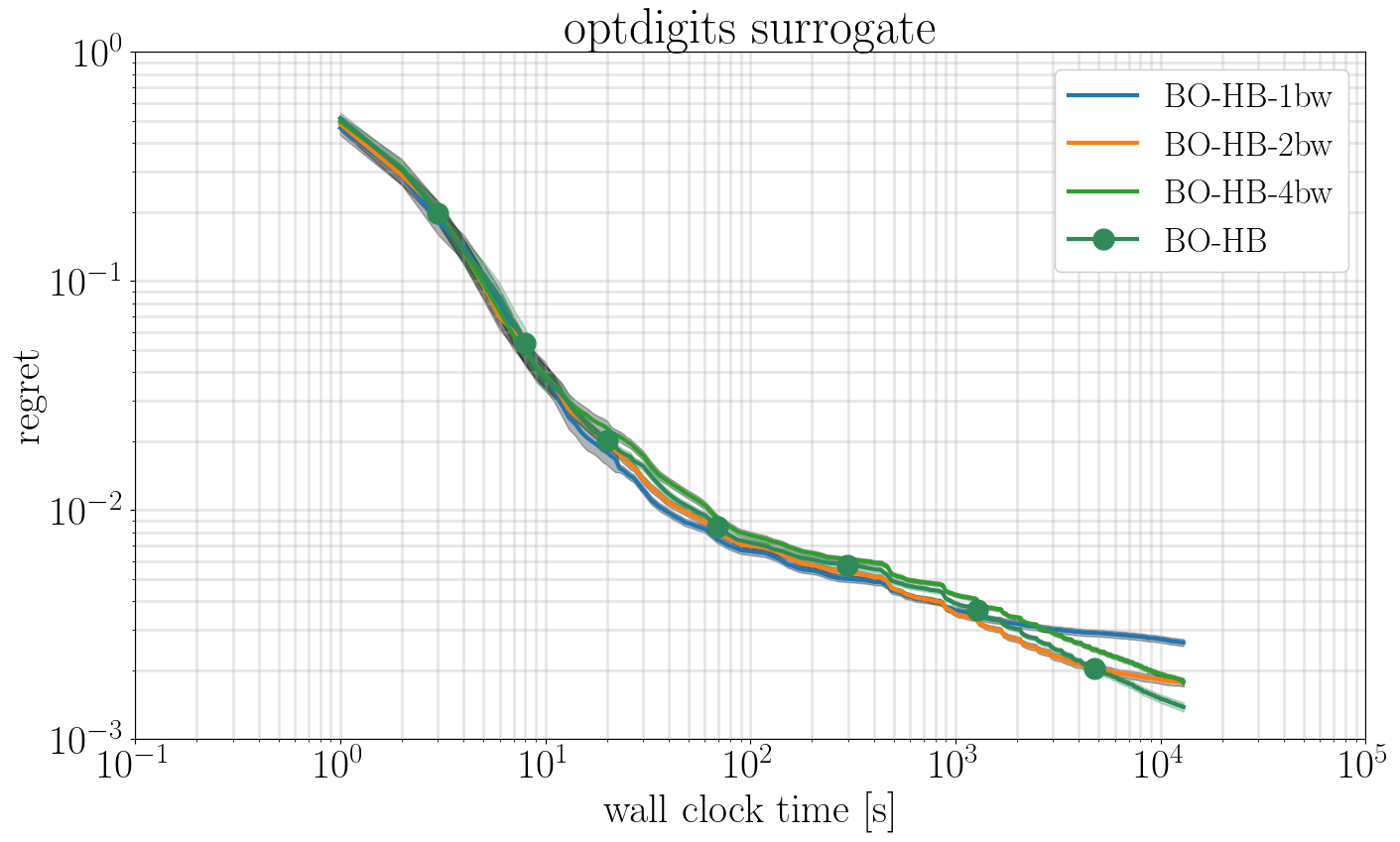}
	\includegraphics[width=0.45\linewidth]{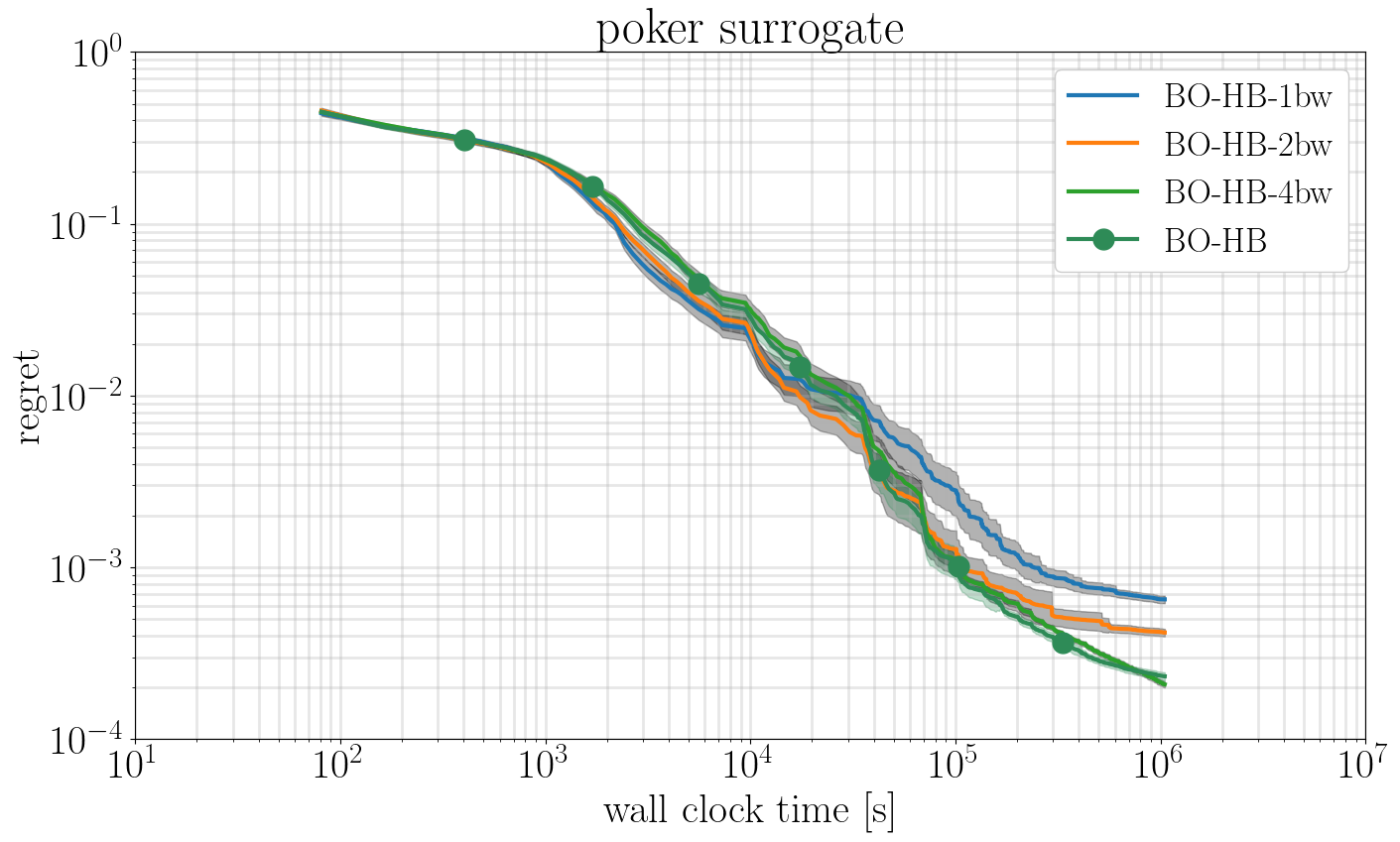}
	\caption{Performance on the surrogates for all six datasets for different bandwidth factors.}
	\label{fig:paramnet_bandwidth_factor_all}
\end{figure*}

In this section, we evaluate the importance of the individual hyperparameters of {\ourmethod}, namely the number of samples used to optimize the acquisition function (Figure \ref{fig:paramnet_num_samples_all}), the fraction of purely random configuration $\rho$ (Figure \ref{fig:paramnet_random_fraction_all}), the scaling parameter $\eta$ (Figure \ref{fig:paramnet_etas_all}), and the bandwidth factor used to encourage exploration (Figure \ref{fig:paramnet_bandwidth_factor_all}).

Additionally, we want to discuss the importance of $\eta$, $b_{min}$ and $b_{max}$ already present in HB.
The parameter $\eta$ controls how aggressively SH cuts down the budget and the number of configurations evaluated.
Like HB \cite{li-iclr17}, {\ourmethod} is also quite insensitive to this choice in a reasonable range.
For our experiments, we use the same default value ($\eta = 3$) for HB and {\ourmethod}.

More important for the optimization are $b_{min}$ and $b_{max}$, which are problem specific and inputs to both HB and {\ourmethod}.
While the maximum budget is often naturally defined, or is constrained by compute resources, the situation for the minimum budget is often different.
To get substantial speedups, an evaluation with a budget of $b_{min}$ should contain some information about the quality of a configuration with larger budgets; for example, when subsampling the data, the smallest subset should not be one datum, but rather enough points to fit a meaningful model.
This requires knowledge about the benchmark and the algorithm being optimized.


\begin{figure*}[t]
	\centering
 \includegraphics[width=\columnwidth]{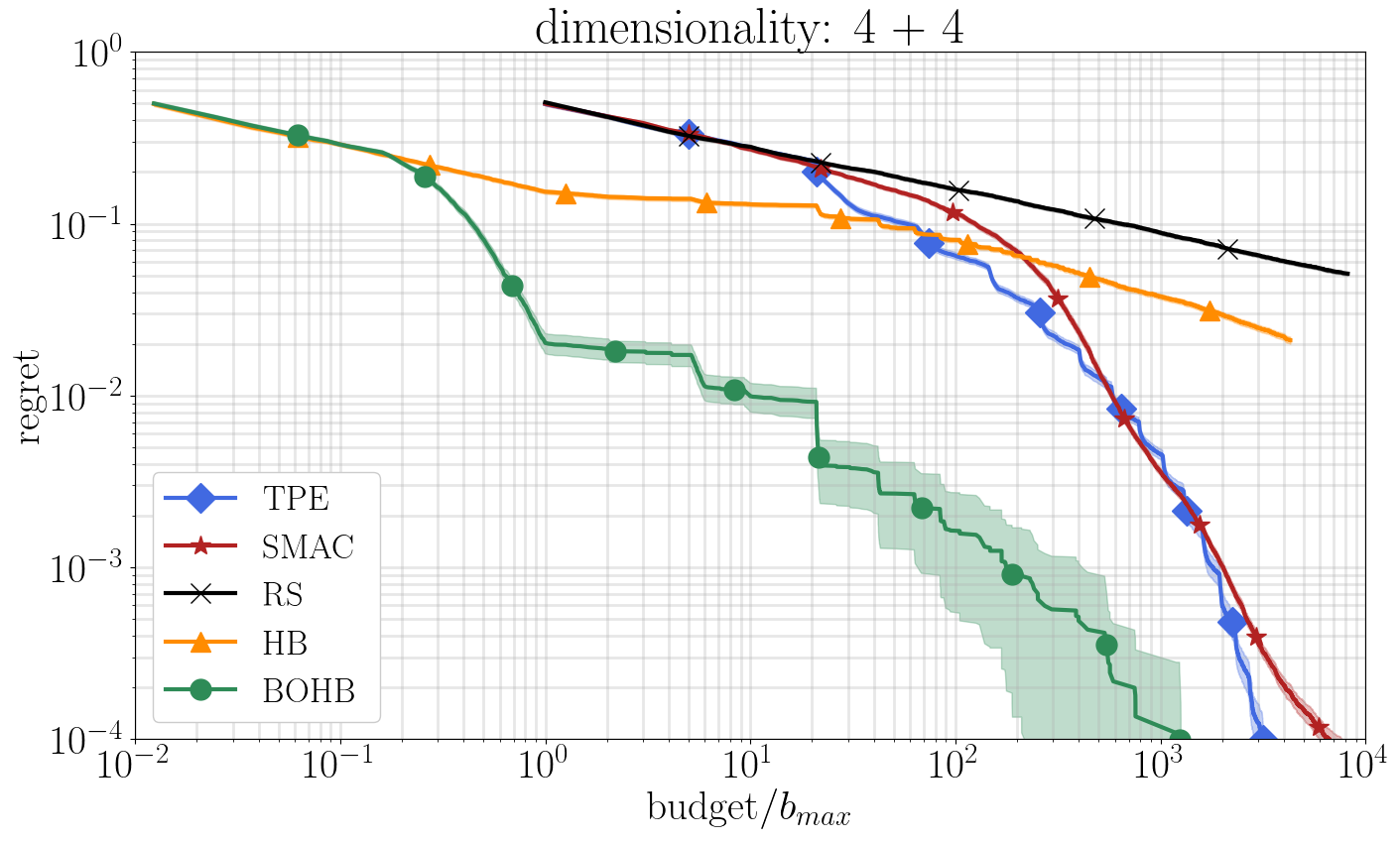}
 \includegraphics[width=\columnwidth]{plots/counting_ones/8_dim_regret.png}
 \includegraphics[width=\columnwidth]{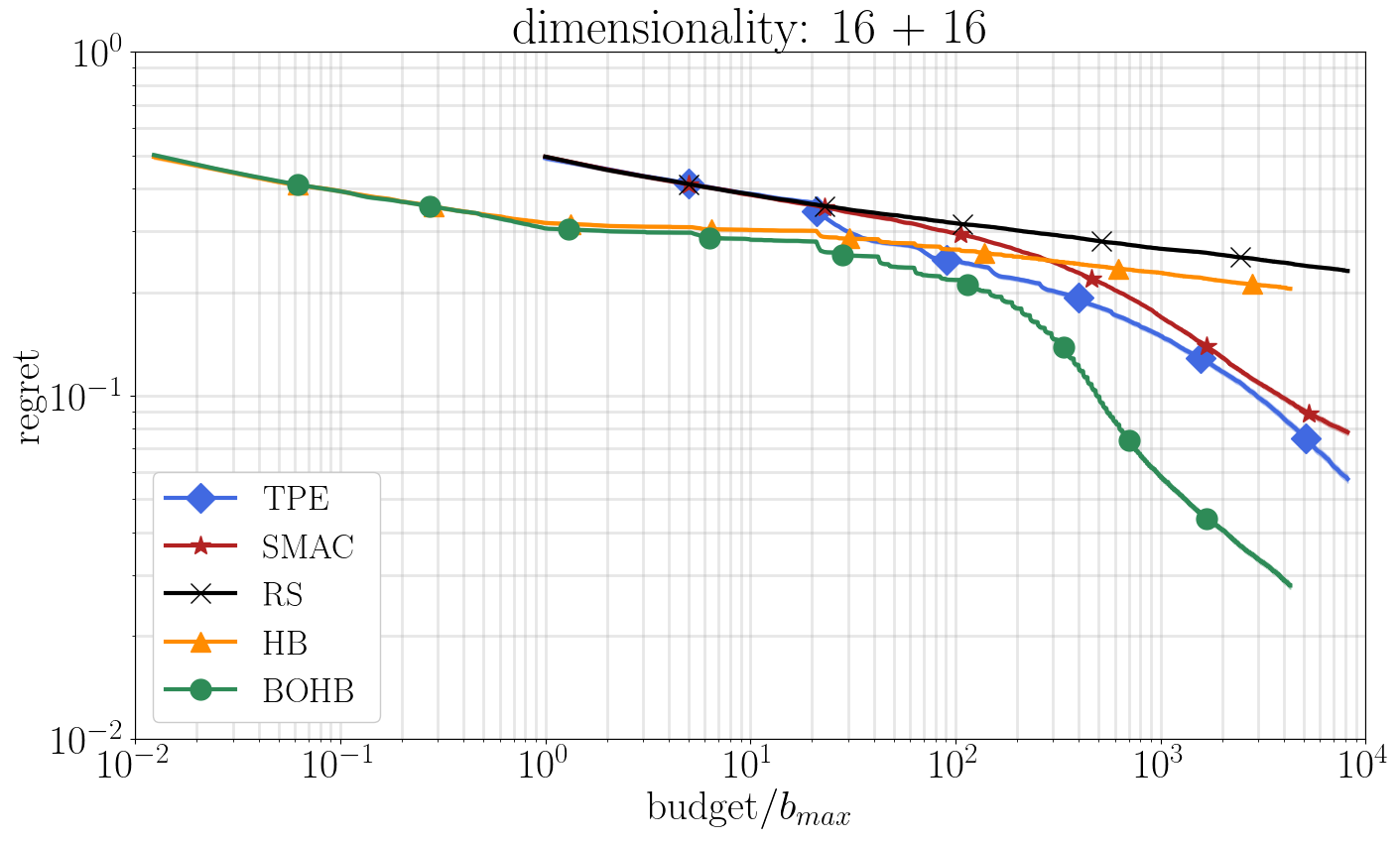}
 \includegraphics[width=\columnwidth]{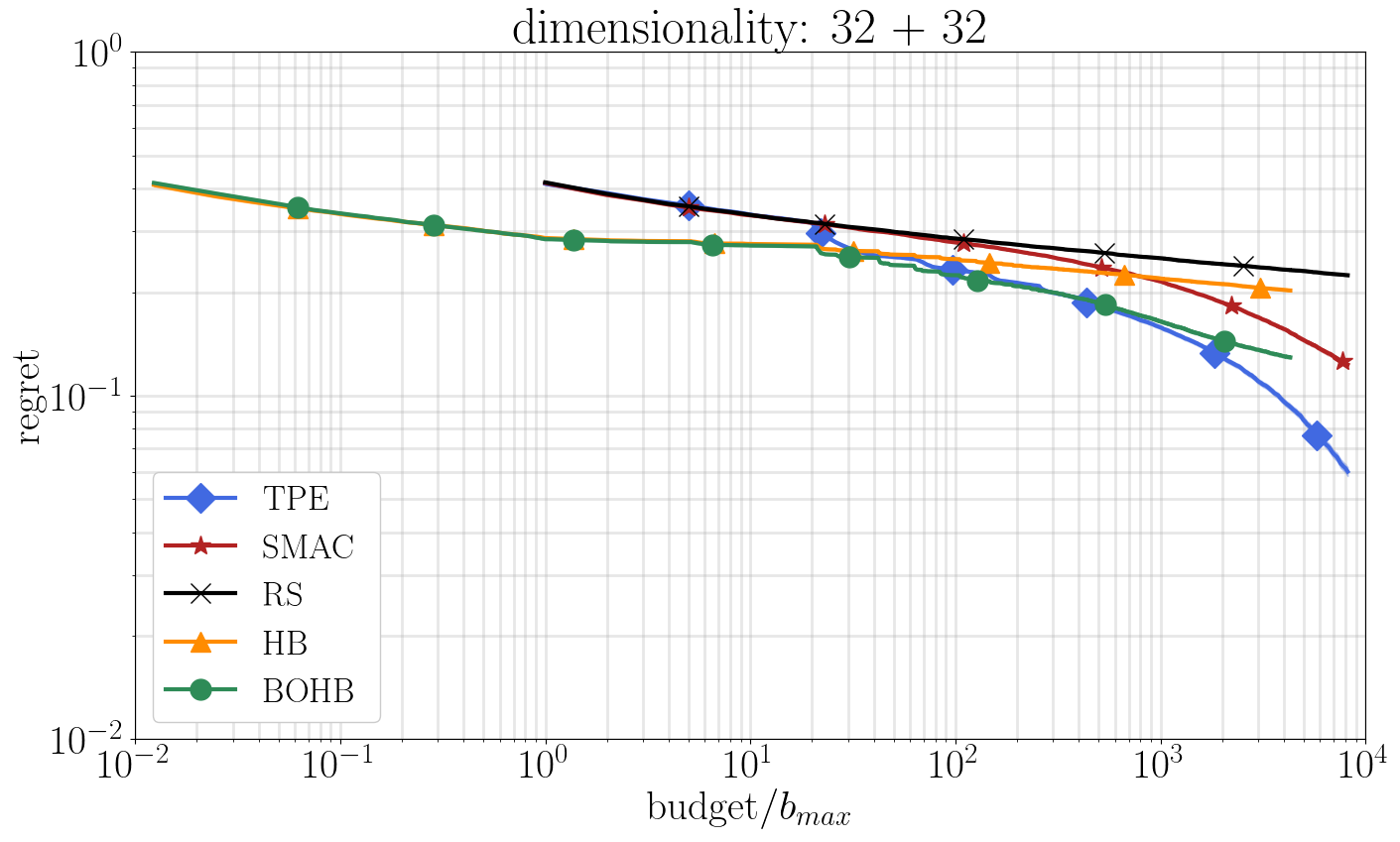}
	\caption{Mean performance of \ourmethod{}, HB, TPE, SMAC and RS on the mixed domain counting ones function with different dimensions. As uncertainties, we show the standard error of the mean based on 512 runs.}
	\label{fig:counting_ones_all_mean}
\end{figure*}

\section{Counting Ones}

We now show results for the counting ones function for different dimensions.
Figure \ref{fig:counting_ones_all_mean} shows the mean performance of all applicable methods in $d$ = 8, 16, 32 and 64 dimensions for a budget of 8192 full function evaluations. 

We draw the following conclusions from the results:
\begin{enumerate}
	\item Despite its simple definition, this problem is quite challenging for the methods we applied to it.
		RS and HB both suffer from the fact that drawing configurations at random performs quite poorly in this space. 
		The model-based approaches SMAC and TPE performed substantially better, especially with large budgets. They required a larger number of samples before converging to the true optimum than \ourmethod{}. However, we would like to mention that SMAC and TPE treated the problem as a blackbox optimization problem; the results for SMAC could likely be improved by treating individual samples as ``instances'' and using SMAC's intensification mechanism to reject poor configurations based on few samples and evaluate promising configurations with more samples.   
	\item {\ourmethod} struggles in the very high dimensional case. We attribute this to the fact that the noise is substantially higher in this case, such that larger budgets are required to build a good model. Therefore, given a large enough budget, \ourmethod{}'s evaluations on small budgets lead to a constant overhead over only using the more reliable evaluations on larger budgets. Since the optimization problem is perfectly separable (there are no interaction effects between any dimensions), we also expect TPE's univariate KDE to perform better than \ourmethod{}'s multivariate one.
\end{enumerate}


\section{Surrogates}

\subsection{Constructing the Surrogates}

To build a surrogate, we sampled 10\,000 random configurations for each dataset, trained them for 50 epochs, and recorded their classification error after each epoch, along with their total training time.
We fitted two independent random forests that predict these two quantities as a function of the hyperparameter configuration used.
This enabled us to predict the classification error as a function of time with sufficient accuracy.
As almost all networks converged within the 50 epochs, we extend the curves by the last obtained value if the budget would allow for more epochs.

The surrogates enable cheap benchmarking, allowing us to run each algorithm 256 times.
Since evaluating a configuration with the random forest is inexpensive, we used a global optimizer (differential evolution) to find the true optimum.
We allowed the optimizer 10\,000 iterations which should be sufficient to find the true optimum.

Besides these positive aspects of benchmarking with surrogates, there are also some drawbacks that we want to mention explicitly:
\begin{enumerate}[(a)]
\item There is no guarantee that the surrogate actually reflects the important properties of the true benchmark.
\item The presented results show the optimized classification error on the validation set used during training. There is no test performance that could indicate overfitting.
\item Training with stochastic gradient descent is an inherently noisy process, i.e. two evaluations of the same configuration can result in different performances. This is not at all reflected by our surrogates, making them a potentially easier to optimize than the true benchmark they are based on.
\item By fixing the budgets (see below) and having deterministic surrogates, the global minima might be the result of some small fluctuations in the classification error in the surrogates' training data. That means that the surrogate's minimizer might not be the true minimizer of the real benchmark. 
\end{enumerate}

None of these downsides necessarily have substantial implications for comparing different optimizers; they simply show that the surrogate benchmarks are not perfect models for the real benchmark they mimic. Nevertheless, we believe that, especially for development of novel algorithms, the positive aspects outweigh the negative ones.

\begin{table}[t]
\centering
\caption{The hyperparameters and architecture choices for the fully connected networks.}
\begin{tabular}{c|c|c}
\hline
	Hyperparameter & Range & Log-transform\\
	\hline
	batch size & $[2^3, 2^8]\rule{0pt}{12pt}$  & yes\\
	dropout rate & $[0,0.5]$ & no\\
	initial learning rate & $[10^{-6}, 10^{-2}]$ & yes\\
	exponential decay factor & $[-0.185, 0]$ & no\\
	\# hidden layers & $\{1, 2,3,4,5\}$ & no\\
	\# units per layer & $[2^{4}, 2^{8}]$ & yes\\
\hline
\end{tabular}
\label{tab:paramnet}
\end{table}

\subsection{Determining the budgets}

To choose the largest budget for training, we looked at the best configuration as predicted by the surrogate and its training time.
We chose the closest power of 3 (because we also used $\eta = 3$ for HB and {\ourmethod}) to achieve that performance.
We chose the smallest budget for HB such that most configurations had finished at least one epoch. Table \ref{tab:budgets} lists the budgets used for all datasets.

\begin{table}[t]
\centering
\vspace*{0.3cm}
\caption{The budgets used by HB and {\ourmethod}; random search and TPE only used the last budget}
\begin{tabular}{c|c}
	Dataset & Budgets in seconds for HB and {\ourmethod}\\
	\hline
	Adult     & 9, 27, 81, 243\\
	Higgs     & 9, 27, 81, 243\\
	Letter    & 3, 9, 27, 81\\
	Poker     & 81, 243, 729, 2187\\

\end{tabular}
\label{tab:budgets}
\end{table}

\section{Bayesian Neural Networks}

\begin{figure*}[t]
	\centering
	\includegraphics[width=0.45\linewidth]{pgf_plots/bnn_boston.pdf}
	\includegraphics[width=0.45\linewidth]{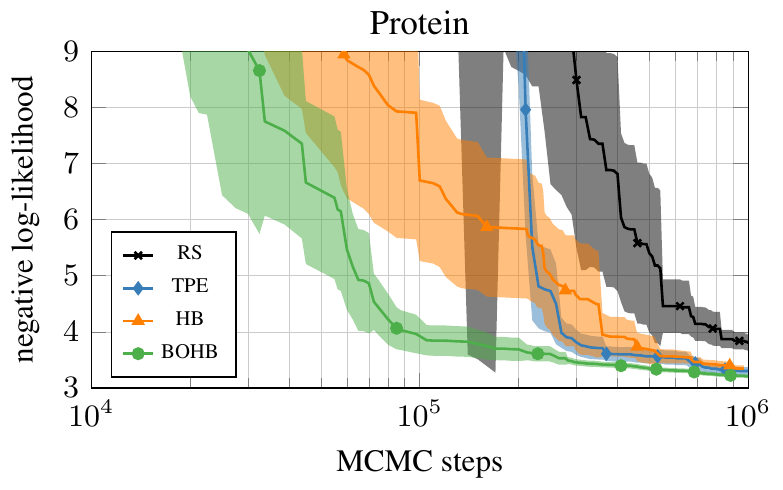}
	\caption{Mean performance of TPE, RS, HB and \ourmethod{} for optimizing the 5 hyperparameters of a Bayesian neural network on two different UCI datasets. As uncertainties, we show the stardard error of the mean based on 50 runs.}
	\label{fig:bnn_all}
\end{figure*}

We optimized the hyperparameters described in Table \ref{tab:bnn} for a Bayesian neural network trained with SGHMC on two UCI regression datasets: Boston Housing and Protein Structure.
The budget for this benchmark was the number of steps for the MCMC sampler. 
We set the minimum budget to 500 steps and the maximum budget to 10000 steps. 
After sampling 100 parameter vectors, we computed the log-likelihood on the validation dataset by averaging the predictive mean and variances of the individual models.
The performance of all methods for both datasets is shown in Figure \ref{fig:bnn_all}.

\begin{table}[t]
\centering
\caption{The hyperparameters for the Bayesian neural network task.}
\begin{tabular}{c|c|c}
\hline
	Hyperparameter & Range & Log-transform\\
	\hline
	\# units layer 1 & $[2^{4}, 2^{9}]$ & yes\\
	\# units layer 2 & $[2^{4}, 2^{9}]$ & yes\\
	step length & $[10^{-6}, 10^{-1}]$ & yes\\
	burn in & $[0, .8]$ & no\\
	momentum decay & $[0, 1]$ & no\\
\hline
\end{tabular}
\label{tab:bnn}
\end{table}

\section{Reinforcement Learning}
Table \ref{tab:cartpole} shows the hyperparameters we optimized for the PPO Cartpole task.

\begin{table}[t]
\centering
\caption{The hyperparameters for the PPO Cartpole task.}
\begin{tabular}{c|c|c}
\hline
	Hyperparameter & Range & Log-transform\\
	\hline
	\# units layer 1 & $[2^{3}, 2^{7}]$ & yes\\
	\# units layer 2 & $[2^{3}, 2^{7}]$ & yes\\
	batch size & $[2^3, 2^8]$  & yes\\
	learning rate & $[10^{-7}, 10^{-1}]$ & yes\\
	discount & $[0, 1]$ & no\\
	likelihood ratio clipping & $[0, 1]$ & no\\
	entropy regularization & $[0, 1]$ & no\\
\hline
\end{tabular}
\label{tab:cartpole}
\end{table}

\bibliography{local,strings,lib,proc}
\bibliographystyle{icml2018}